\documentclass[review]{elsarticle}

\usepackage{lineno,hyperref}
\usepackage{array}
\usepackage{booktabs}
\usepackage{mathtools}
\usepackage{multirow}
\usepackage{changepage}
\usepackage{textcomp,marvosym}
\usepackage{microtype}
\DisableLigatures[f]{encoding = *, family = * }

\usepackage[table]{xcolor}

\usepackage[utf8x]{inputenc}
\usepackage{caption}
\usepackage{subcaption}
\usepackage{amsmath,amssymb}

\newlength\savedwidth

\newcommand\thickhline{\noalign{\global\savedwidth\arrayrulewidth\global\arrayrulewidth 2pt}%
\hline
\noalign{\global\arrayrulewidth\savedwidth}}
\usepackage{wrapfig}
\usepackage[euler]{textgreek}

\usepackage[aboveskip=1pt,labelfont=bf,labelsep=period,justification=raggedright,singlelinecheck=off]{caption}

\makeatletter
\def\els@aparagraph[#1]#2{\elsparagraph[#1]{#2}}
\def\els@bparagraph#1{\elsparagraph*{#1}}
\makeatother

\modulolinenumbers[5]

\makeatletter
\def\ps@pprintTitle{%
   \let\@oddhead\@empty
   \let\@evenhead\@empty
   \let\@oddfoot\@empty
   \let\@evenfoot\@oddfoot
}
\makeatother

\biboptions{authoryear}

\begin{document}

\begin{frontmatter}

\title{Hierarchical Deep Multi-modal Network for Medical Visual Question Answering}
%  \tnotetext[mytitlenote]{Fully documented templates are available in the elsarticle package on \href{http://www.ctan.org/tex-archive/macros/latex/contrib/elsarticle}{CTAN}.}

    \author{Deepak Gupta\corref{cor-2}\fnref{fn1}\corref{cor1}}
    \ead{deepak.pcs16@iitp.ac.in}
  \author{Swati Suman\fnref{fn2}}
  \ead{swati17293@gmail.com}
     \cortext[cor1]{Corresponding author}
     \fntext[fn2]{Both authors contributed equally to this work.}
    \author{Asif Ekbal\corref{cor2}}
 \ead{asif@iitp.ac.in}
    \address{Department of Computer Science and Engineering\\
    Indian Institute of Technology Patna, India}
    
% \author{Swati, \corref{cor1}Deepak Gupta, Asif Ekbal}
% \address{Department of Computer Science and Engineering}
%  \cortext[cor1]{Corresponding author}
%  \ead{swati@elsevier.com}

% \author[mymainaddress,mysecondaryaddress]{Elsevier Inc}

% \author[mysecondaryaddress]{Global Customer Service\corref{mycorrespondingauthor}}
% \cortext[mycorrespondingauthor]{Corresponding author}

% \address[mymainaddress]{1600 John F Kennedy Boulevard, Philadelphia}
% \address[mysecondaryaddress]{360 Park Avenue South, New York}

\begin{abstract}
       {
       Visual Question Answering in Medical domain (VQA-Med) plays an important role in providing medical assistance to the end-users. These users are expected to raise either a straightforward question with a \textit{Yes/No} answer or a challenging question that requires a detailed and descriptive answer. The existing techniques in VQA-Med fail to distinguish between the different question types sometimes complicates the simpler problems, or over-simplifies the complicated ones. It is certainly true that for different question types, several distinct systems can lead to confusion and discomfort for the end-users. To address this issue, we propose a hierarchical deep multi-modal network that analyzes and classifies end-user questions/queries and then incorporates a query-specific approach for answer prediction. We refer our proposed approach as \textbf{H}ierarchical \textbf{Q}uestion \textbf{S}egregation based \textbf{V}isual \textbf{Q}uestion \textbf{A}nswering, in short HQS-VQA. 
    %   We first use the Support Vector Machine (SVM) with the hand-engineered features to classify the questions into \textit{yes/no} and \textit{descriptive} types.
   % Then, based on the question types, we employ different strategies to provide the answer. The \textit{Yes/No} type questions are treated as a binary classification problem. We generate the answer from a fixed vocabulary for the \textit{descriptive} type question. 
   Our contributions are three-fold, \textit{viz.} firstly, we propose a question segregation (QS) technique for VQA-Med; secondly, we integrate the QS model to the hierarchical deep multi-modal neural network to generate proper answers to the queries related to medical images; and thirdly, we study the impact of QS in Medical-VQA by comparing the performance of the proposed model with QS and a model without QS. We evaluate the performance of our proposed model on two benchmark datasets, \textit{viz.} RAD and CLEF18. Experimental results show that our proposed HQS-VQA technique outperforms the baseline models with significant margins. We also conduct a detailed quantitative and qualitative analysis of the obtained results and discover potential causes of errors and their solutions.
        }
\end{abstract}


\end{frontmatter}
\setlength{\parskip}{1em}
% \linenumbers

\section{Introduction}
\label{sec:introduction}
        The advancement in the field of Computer Vision (CV)~\citep{arai2019advances,guo2020gluoncv,li2020transfer} and Natural Language Processing (NLP)~\citep{ruder2019neural,ruder2019transfer,fu2019introduction,dong2019unified} over the last decade, has introduced several interesting machine learning techniques. %problems more convenient. 
        The problems such as object detection~\citep{liu2020deep}, segmentation~\citep{liu2019auto}, and  image classification~\citep{sun2020few,sun2020automatically} in CV, and machine translation~\citep{yang2020hierarchical}, question answering~\citep{gupta2018uncovering,gupta2019deep,chen2016thorough,mmqa,C18-1042}, biomedical and clinical text mining~ \citep{ningthoujam2019relation, yadav2018multi,yadav2019unified,yadav2020exploring,chen2019constructing}, speech recognition~\citep{magnuson2020earshot} in NLP, are being solved much more efficiently than ever before. This has facilitated the researchers to indulge into solving interdisciplinary problems that demand knowledge of both the fields. 
        
        Visual Question Answering (VQA)~\citep{gao2019dynamic,kafle2020answering} has emerged as one such problem. In VQA, the task is poised as questions being asked with respect to an image, where the machine needs to learn and generate answers of such questions based on the learned features of the input image. In contrast to the typical CV tasks which largely focus on %have singular and 
        solving problems such as %restricted problems (eg., 
        \textit{action identification}, \textit{image classification}, VQA tasks are relatively complex. It demands more intelligence like object recognition, semantic feature extraction, external knowledge, and common sense knowledge. Many domain-specific VQA tasks have surfaced in the last few years, and VQA in the medical domain is one such that plays a significant role in providing medical assistance. Since this task is related to the medical domain, end-users can be categorized based on the type of queries raised by them. Patients, medical students, and related users are anticipated to ask elementary questions mostly having \textit{Yes/No} as the answer. On the other hand, clinicians and medical experts are expected to raise a more problem-specific query demanding a detailed and descriptive answer. For this reason, different portals must be created to satisfy the query-specific need, but that would lead to confusion and discomfort to end-users. To tackle this problem, in this paper, we propose a question segregation technique to segregate the user queries. For this module, we use a simple statistical machine learning model, based on simple hand-engineered, and word frequency-based features. 
        % We then use this technique to implement a hierarchical deep multi-modal network that analyzes and classifies the queries at the root of the hierarchy and then incorporates the query-specific approach at leaves. To predict the answers in this model, we compute the question and image representations, which we first fuse and then pass to the specific answer prediction model at the leaf node. We treat the model as a two-class classification system for questions with type `\textit{Yes/No}' while we treat the model as a multi-label classification model for questions with type `\textit{others}'. We perform the experiments with two different datasets in the medical domain showing that, apart from unifying the two query-specific models, the hierarchical model also enhances the performance of the same model without such hierarchy. It also outperforms the baseline models with significant margins. 
        
         Towards the solution of the problem of generic VQA-Med, we propose a hierarchical deep multi-modal network that analyzes and classifies the queries at the root of the hierarchy and then incorporates the query-specific approach at leaf nodes. To predict the answers in this model, we generate the question and image representations using Bidirectional Long Short Term Memory (Bi-LSTM)~\citep{Graves2005} and Inception-Resnet-v2~\citep{szegedy2017inception}, respectively. We fuse the representations together and pass it to the specific answer prediction model at the leaf node. For the task of question classification in the root node, we propose a question segregation technique. We use Support Vector Machine (SVM)~\citep{cristianini2000introduction} as the classifier with hand-engineered and word frequency-based features for QS. We use the machine learning technique for QS, as the rule-based strategy suffers from the problem of defining too many rules that may not extend to other datasets~\citep{clark2018question}. The following examples from RAdiology Data (RAD)~\citep{lau2018dataset} show the difficulty of the rule-based approach in the medical domain.

        \begin{itemize}
            \item {\bf Question:} Evidence of hemorrhage in the kidneys? \\ {\bf Answer:} No\\
            {\bf Question-type:} \textit{`Yes/No'}
            \item {\bf Question:} Is the spleen present?  \\ {\bf Answer:} on the patient's left\\
            {\bf Question-type:} \textit{`Others'}
        \end{itemize}

        Careful analysis of the question reveals that the first example expects a descriptive type answer that is to list out the facts that indicate kidney hemorrhage (Question-type: \textit{`Others'}), while the second example expects to confirm the presence/absence of spleen (Question-type: \textit{`Yes/No'}). The presence of such anomalies in the question acts as a hindrance in the formation of robust rules for the classification of questions into their correct type.
        
        We perform all our experiments in the RAD and  ImageCLEF2018 VQA-Med 2018 (CLEF18) datasets, as they perfectly capture the problem statement that we intend to solve. Detailed discussion on the dataset can be found in Section~\ref{sec:data-description}. Experimental evaluation demonstrates promising results, showing the effectiveness of our proposed approach. %'s efficiency. 
        Additionally, error analysis of the system's outputs %error analysis 
        shows the future direction in this research area by addressing different kinds of errors.

        The organization of this paper is as follows. We first discuss the related work in VQA. Then we present the details of the methodologies that we implemented to solve our specific problem. In particular, we explain our proposed HQS-VQA models in detail. Basically, we discussed the technique used for the question segregation module and the VQA components used to generate the query-specific answers. Details of experiments along with the evaluation results and necessary analysis are reported. %We then perform the experiments and show the results with qualitative and quantitative analysis. 
        Finally, we conclude and provide the future directions of our work.

    \subsection{Motivation}
    \label{sec:motivation}
    The motivation behind our work are stemmed from the following facts: %of the medical visual question answering are listed as follows:
   \begin{itemize}
       \item 
        Nowadays, medical and physiological images (``{ct-scan}", ``{x-ray}", etc.) and reports for the patients are easily accessible with the increase in the use of medical portals. But the patient still needs to visit a medical expert to fully understand those reports and get answers to their queries. This process is both time-consuming and cost-sensitive. %involves investment of time and money, even to get simple queries answered. 
        On the other hand, clinicians also find an efficient VQA system very useful to understand the results of a complex medical image. They may use such a system as a second opinion just to boost their self-confidence in the understanding of some specific aspects of such medical images. Although it is possible to search queries on search engines, the search results may be inaccurate, spurious, vague, and, or enormous. In the case of medical reports such inaccurate, spurious, or vague results could lead to serious after-effects. In this context, the VQA in the medical domain is getting attention as an important research problem trying to provide the answer to end-user queries related to medical images.
  \item
        VQA-Med intends to assist patients and clinicians in general, but can also be useful in medical education. A clinical apprentice or medical students who just started learning the basics of handling images of different modalities, may learn by asking queries and getting answers. Thus, developing an efficient and automated VQA system for the medical domain comes out as an essential task. Even though many medical datasets are published publicly, most of them deal with some specific disease in a particular body part with a fixed image modality. ImageCLEFtuberculosis task~\citep{cid2018overview} is one such example which was published to build models for detection, classification, and severity measurement of TB from the provided chest-CT. On the contrary, it is a more challenging problem to have a generic strategy that deals with VQA queries regarding multiple image modalities linked to multiple diseases that may appear in any part of the body. The solution to such a generic problem will be considerably less confusing to the patients. 
 \item
        The difference in the type of end-users results in different query types. The queries from patients most of the time are expected to be generic in nature, having the need to produce `\textit{Yes}' or `\textit{No}' as the answer. Whereas, the queries from clinicians and medical experts are expected to be more problem-specific, which requires elaborate answers. Again, a skilled trainee is expected to ask more specific and sophisticated questions, while queries from beginners are likely to be simple and straightforward. For example, a naive trainee may inquire about the presence of any abnormality in the image, whereas a senior trainee may identify the abnormality of `\textit{intraventricular hemorrhage}' from the image, and want to understand more about the grade and effect of the hemorrhage. They can then draw inferences from the acquired data for effective treatment.
\item
        This difference in query type thus needs different problem-specific attention, which needs to be dealt with isolation. Again, multiple end systems for multiple types of queries may create confusion, and discomfort to the end-user. There should be a single end-user module to solve both the complex and simple queries. Table~\ref{table:question-answer-pairs} demonstrates one such system where any clinically relevant question can be asked about the image. Here, image plays an important role as the answer to the questions may vary based on the provided image.
        
        \begin{table}[!ht]
            \begin{adjustwidth}{-0.9in}{-1.2in}
            \centering
            \begin{tabular}{m{0.25\linewidth}  m{0.45\linewidth}  m{0.2\linewidth}}
                {\bf Image} & {\bf Question} & {\bf Answer} \\ \hline
                \multirow{ 4}{*}{\includegraphics[width=0.9\linewidth,height=0.7\linewidth]{images/ex-img-RAD.jpg}} & Is this a cyst in the left lung? &  No \\ \cline{2-3}
                & Has the left lung collapsed? & Yes \\ \cline{2-3}
                & Where is the nodule? & Below the 7th rib in the right lung \\ \cline{2-3}
                & What are the densities in both mid-lung fields? & pleural plaques
                \\ \hline
            \end{tabular}
            \caption{Sample, question-answer pairs formulated from a single image. More than one clinically relevant questions can be asked from a given image. }
            \label{table:question-answer-pairs}
            \end{adjustwidth}
        \end{table}
\item
        We identify this need, and propose a  SVM-based question segregation technique to segregate the questions. We then use this information to propose a hierarchical deep multi-modal network to generate the answers.
  \end{itemize}
    \subsection{Contributions}
    \label{sec:contributions}
            \begin{itemize}
                \item We propose a SVM based Question Segregation technique for the task of question classification for VQA in the medical domain.
                \item We propose a hierarchical deep multi-modal neural model and integrate it with the proposed question segregation module to generate proper answers to the queries related to medical images.
                \item We study the impact of QS in Medical-VQA by comparing the performance of the proposed model with question segregation and a model without such segregation. We also compare it with the baseline models to study its effectiveness.
                \item We evaluate our model on two different datasets,% to study the robustness of our proposed approach,
                which demonstrate the fact that our proposed method is generic in nature. 

            \end{itemize}
            
\section{Related Work}
\label{sec:related-work}
   
        The major challenges of the VQA-Med are closely related to the general VQA and QA in the medical domain and we see a lot of interesting solutions evolving over time. We present the survey with respect to the related datasets and methods in the following subsections.

 \subsection{Datasets}
    \label{sec:medical-datasets}
        A number of research projects have been initiated for the development of benchmark datasets to promote the works in the medical domain. The Genomic corpus released as part of the TREC~\citep{Hersh2003TRECGT} task is one of the benchmark datasets for the medical QA task. It focuses exclusively on scientific papers. However, a small number of questions in the dataset are not sufficient to evaluate the efficiency of the large-scale QA systems. This constraint led to the release of several other datasets, such as Question Answering for Machine Reading Evaluation (QA4MRE)~\citep{articleMorante} and Biomedical Semantic Indexing and Question Answering (BioASQ)~\citep{TsatsaronisGeorge}. The QA4MRE consists of the Biomedical Text on Alzheimer's Disease, while BioASQ gathers information from various heterogeneous sources to address real-life questions from the biomedical experts. A number of datasets, such as MRI-DIR~\citep{Rachel2018}, fastMRI~\citep{fastMRI}, and a few more ~\citep{Erickson2017,Martin2017,Bakr2017} focused on different medical tasks, are also available.
        However, the images in the VQA-Med dataset have different modalities and contain radiological markings such as short information, tags, etc. It may also contain a stack of sub-images that is not the case with the existing medical datasets. In addition, general VQA datasets~\citep{lin2014microsoft,mukuze2018vision,gebhardt2018camel,antol2015vqa} are task-specific, unlike VQA-Med, where a question can be asked about any disease from any part of the body.
    
        % MS-COCO~\citep{lin2014microsoft} is the most popular dataset that contains real world images along with their corresponding captions. Most of the VQA datasets~\citep{mukuze2018vision,gebhardt2018camel} were created using the image captions from MS-COCO. Visual Question Answering (VQA 1.0)~\citep{antol2015vqa} is one of them. It is the most significant and commonly used dataset for the VQA task which was published as a part of the VQA challenge. The VQA 1.0 dataset's primary issue was its inherent bias and language priors too had a major impact on the responses to the questions that inspired the design of the second version of this dataset. VQA 2.0~\citep{malinowski2015ask} is a larger and more balanced dataset. 
   
        In this work, we use the RAD ~\citep{lau2018dataset} and CLEF18 ~\citep{ionescu2018overview} medical VQA datasets, which are  different from the existing VQA datasets. The obvious reason is their focus on the medical domain, which offers distinguishing challenges. The images, questions, and answers must be clinically relevant in order to be a part of this dataset which is not a constraint in the VQA datasets. 
        % The building of phrases for sentences is another distinction. Most of the Medical-VQA sentences are complex with a lot of medical terms while being simple and straightforward in the stated datasets. Another difference is the incomparable size of the dataset, medical domain data resources are limited compared to the general domain data resources which are usually huge (e.g., thousands vs. millions of samples). The number of reference answers, which is just one, is another drawback of this dataset.

    \subsection{Methods}
    \label{methods}
   
\begin{figure}[!ht]
            \centering
            \includegraphics[width=1.0\textwidth,height=0.41\textwidth]{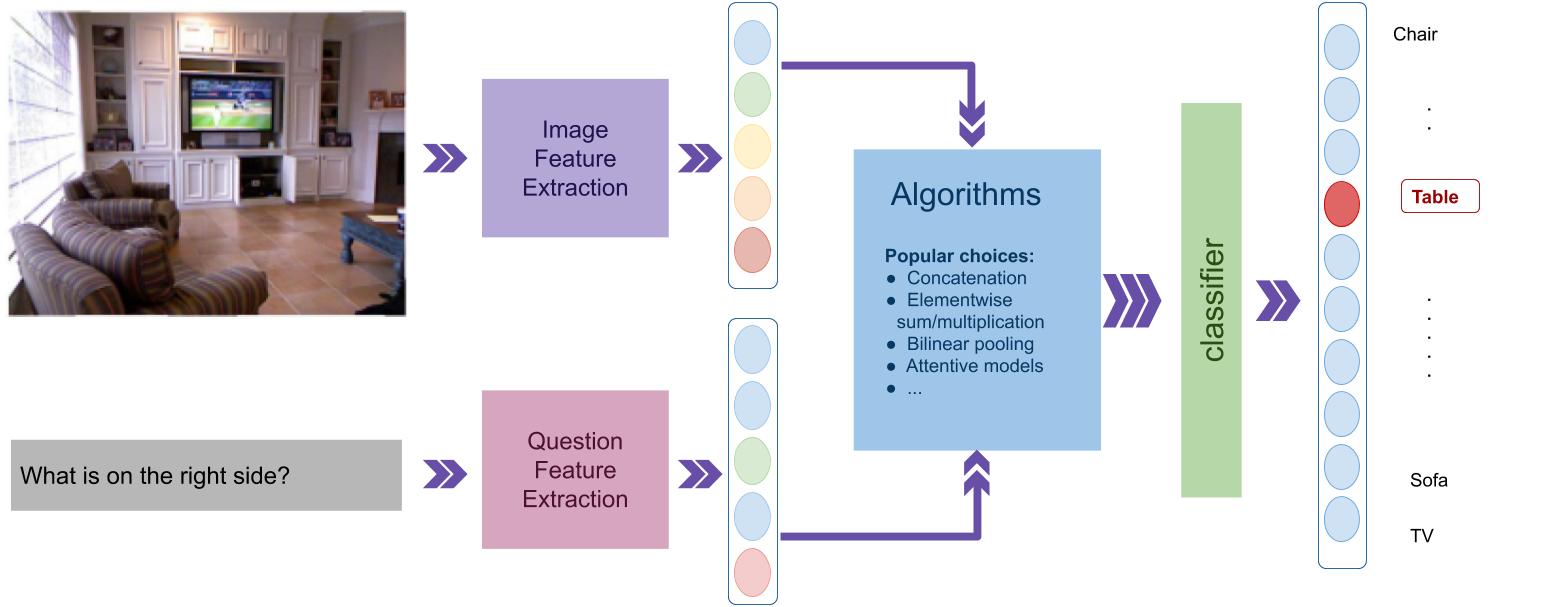}
            \caption{Framework for VQA where, question and image are taken as input  to generate or predict the answer.}
            \label{fig:key-component-framework}
        \end{figure}

        VQA tasks are primarily based on three key components: generating representations of images and questions; passing these inputs through a neural network to produce a co-dependent embedding, and then generating the correct response. Fig~\ref{fig:key-component-framework} illustrates this framework where the key components can take a wide variety of forms. 

        VQA systems differ from each other in the way they fuse multi-modal information. Although most open-ended VQA algorithms used the classification mechanism, this strategy can only produce answers seen during training. The multi-word response is generated one word at a time using LSTM~\citep{Gao:2015:YTM:2969442.2969496, malinowski2015ask}. The response generated, however, is still restricted to words seen in the course of training.

        For question encoding, most methods for VQA uses a variant of Recurrent Neural Network (RNN)~\citep{mikolov2010recurrent}. RNNs are capable of handling sequence problems, but when RNN processes lengthy sequences, context data is easily ignored. LSTM's ~\citep{greff2016lstm} proposal mitigated the long-distance dependency issue. In addition, the researchers also discovered that the respective route from the decoder to the encoder will be reduced if the input sequence is inverted, contributing to network memory. The Bi-LSTM~\citep{Graves2005} model combines the above two points and improves the results. The Gated Recurrent Unit (GRU)~\citep{cho-etal-2014-learning} is also notable, and widely used, simplification of the LSTM. As for the image feature extraction, Convolutional Neural Network (CNN)~\citep{traore2018deep} are used where VGG-net~\citep{Simonyan15} and deep residual networks (ResNet)~\citep{he2016deep} are the most popular choice.  
   
        The application of attention on the image can help to improve the performance of the model by discarding the irrelevant parts of the image. So, attention mechanisms~\citep{Xiong:2016:DMN:3045390.3045643, yang2016stacked} are usually incorporated in the models so that they may learn to attend to the important regions of the input image. However, attending image is not enough but question attention is important too as most of the words in the question may be irrelevant so simultaneous integration of both question and image attention is advised~\citep{lu2016hierarchical}. The fundamental concept behind all these attentive models is that for answering a specific question, certain visual areas in an image and certain words in a question provides more information than others. The Stacked Attention Network (SAN)~\citep{yang2016stacked} and the Dynamic Memory Network (DMN)~\citep{Xiong:2016:DMN:3045390.3045643} used image features from a CNN feature map's spatial grid. In~\citep{yang2016stacked} an attention layer is specified by a single layer of weights using the question and image feature defined to calculate attention distribution across image locations. Using a weighted sum, this allocation is then applied to the CNN feature map to pool across spatial feature locations. It creates a global representation of the image that highlights certain spatial regions.

        VQA depends on the image and question being processed together. This was achieved earlier by using simplified methods such as concatenation or element-wise product, but these methods fail to capture the complex interactions between these two modalities. Later, multi-modal bi-linear pooling was proposed where the idea was to approximate the outer product between the two features, enabling a much deeper interaction between them. Similar concepts have been shown to work well to improve the fine-grained image recognition~\citep{lin2017bilinear}. Multimodal Compact Bilinear (MCB)~\citep{fukui-etal-2016-multimodal} is the most significant VQA technique used in bilinear pooling. It calculates the outer product in a reduced dimensional space instead of explicit calculation to minimize the number of parameters to be learned. Then this is used to predict the relevant spatial features according to the question. The major change was the use of MCB for feature fusion instead of element-wise multiplication.

        Methods for Medical-VQA must be different from general VQA as the size of the datasets is incomparable. The other challenge with Medical-VQA is to balance the number of image features (usually thousands) with the number of clinical features (usually just a few) in the deep learning network to avoid drowning out of the clinical features. Attention-based on bounding box too cannot be applied directly as medical images lack the bounding box information. For medical imaging, there are many computer-aided diagnostic systems~\citep{Kawahara2016MultiresolutionTractCW,7401039,tarando2016increasing}. Most of them, however, deal with single disease problems and focused primarily on easily identifiable areas such as the lungs and skin. In contrast to these systems, Medical-VQA deals with multiple diseases at the same time apart from handling multiple body parts which are difficult for machines to learn.
    
 {Recently, the ImageCLEF introduced the challenge of Medical Domain Visual Question Answering, VQA-Med 2018\footnote{\url{https://www.imageclef.org/2018/VQA-Med}} \citep{ionescu2018overview}. The system submitted by \cite{peng2018umass} achieved the best performance (in terms of BLEU score) in VQA-Med 2018 for medical visual question answering. They built their best performing systems using ResNet-152 for image feature extraction and Multimodal Factorized High-order (MFH) \cite{yu2018beyond} for language-vision fusion. \citet{zhou2018employing} utilized the Inception-Resnet-v2 and Bi-LSTM for image and question representation, respectively. They used the inter-attention mechanism to fuse the language and vision features. Their best performing system stood second among all the submitted systems in the challenge. The third best system submitted by \citet{abacha2018nlm} uses the pre-trained VGG-16 model for image feature extraction and LSTM for question representation. They utilized the stack attention network to fuse the question and image features. In the second edition\footnote{\url{https://www.imageclef.org/2019/medical/vqa/}} of VQA-Med, \citet{yan2019zhejiang} submitted the best system for medical visual question answering. The proposed approach utilized the BERT \citep{devlin2018bert} for question representation and pre-trained VGG-16 model for image representation. They fused the question and image features using MFB mechanism.
    }
    % here were only limited datasets ($5413$ train, $500$ validation and $500$ test) were released as part of the VQA-Med 2018 and challenge received a total of $17$ runs from $5$ participating teams. In the second edition of the VQA-Med 2019\footnote{}  
    
        Inherently, questions follow a temporal sequence and naturally cluster into different types. This question-type information is very important to predict the response regardless of the image. Authors in~\citep{7780907} use a similar approach where they first identify the question-type and use this information for answer generation. Our work, however, isolates the learning path based on question-type rather than using this knowledge as a feature. This type of information can also affect model performance as some of the VQA models perform better than others for certain types of questions. Therefore, these models can be intelligently combined to leverage their varied strengths. We propose a simple model with a question segregation module which segregates the learning path based on the question types (\textit{Yes/No} and \textit{Others}) to reap the benefits of question-type dedicated models. We use Inception-Resnet to encode image feature and Bi-LSTM for question feature creation. 

\section{Materials and Methods}
\label{sec:materials-and-methods}
    \subsection{Problem Modeling}
   
        Given a pair $(Q, I)$, where $Q$ is the textual question accompanied by any medically relevant image $I$, the Medical-VQA task is aimed to generate the appropriate answer $A$. Mathematically, it can be formulated as,
        \begin{eqnarray}
        \label{eq:Main-Problem}
	        A = f(Q,I, \alpha)
        \end{eqnarray}
        where $f$ is the answer prediction function and $\alpha$ denotes the model parameters. Questions in $Q$ can be categorized into two question types ($q\_type$). For questions with $q\_type=Yes/No$, the input $Q$ have a straightforward binary response. While for questions with $q\_type=Others$, it can have a well-thought-out variable length response generated from the answer dictionary words. The problem is to develop a hierarchical model with a question segregation module to differentiate the learning path for the two $q\_type$ for solving the generic Medical-VQA task.

    \subsection{Methodology}
    \label{sec:methodology}
   
\begin{figure}[!ht]
            \includegraphics[width=0.75\textwidth,height=0.2\textwidth]{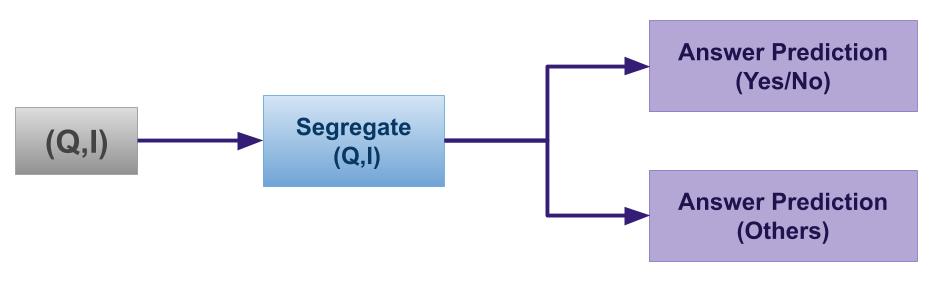}
            \caption{Abstract representation of the proposed hierarchical HQS-VQA model. The first level segregates the questions while the second level generates the answer using the leaf node. Answer prediction strategy is decided based on the question type.}
            \label{fig:hierarchy}
        \end{figure}

      Our proposed approach towards the solution of the problem is to form a two-level hierarchy, where the top-level task is question segregation and the next-level task is answer prediction. The proposed hierarchical model is depicted in Fig~\ref{fig:hierarchy}. The subsequent sections describe the components of the proposed hierarchy.

    \subsubsection{Question Segregation}
    \label{sec:question-segregation}
   
\begin{figure}[!ht]
.           \includegraphics[width=0.85\textwidth,height=0.22\textwidth]{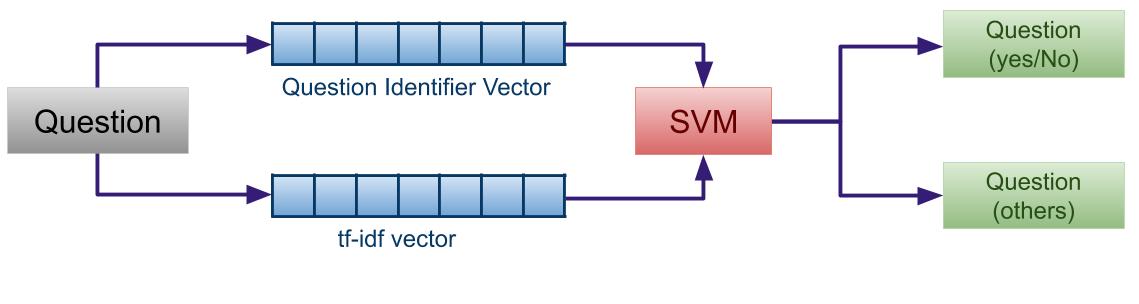}
            \caption{Proposed question segregation module with linear SVM learner as base classifier. The extracted feature vectors are fed to the SVM for question segregation.}
            \label{fig:qs-module}
        \end{figure}

        Question Segregation, in general, segregates the questions from a question list $Q = [q_{1}, \;  q_{2} \ldots \; q_{n}]$ based on the $q\_type$, where $n$ denotes the total number of questions. In this list, $q_{i} = ``w_{1} \; w_{2} \ldots \; w_{m}"$ denotes the $i^{th}$ question in the list containing a sequence of $m$ words. The task of question segregation is relatively an easier task compared to answer prediction. Thus, we find a simple statistical machine learning model based on simple hand-engineered, and word frequency-based features to effectively solve the problem poised in the top-tier (segregation) of our proposed hierarchical setup. We employ an SVM classifier for this purpose. %We used question classifier to segregate the question into Yes/No or descriptive types. 
        The SVM is relatively less complex compared to the deep neural network.  Moreover, the datasets, which we use here are relatively smaller, and hence deep learning models tend to perform on the lower-side compared to the classical supervised SVM based model. 
        % small and the questions of both the types are unevenly distributed, resulting in major class imbalances. In such cases, the deep learning network tends to over-fit while the SVMs are less likely to over-fit and perform better as depicted in the results.} 
        Fig~\ref{fig:qs-module} illustrates the entire question segregation process. 

    The classifier, and input to QS module i.e. question feature vectors generated from the questions in the dataset are explained as follows:

    \paragraph{{\bf Question Feature Vectors:}} 
    \label{sec:question-feature-vectors}
        From each question, we extract the following two vectors:
        \begin{itemize}
            \item \textbf{Question Identifier Vector:} We form a set of $r$ question identifier words, where each word tries to represent the question motive. We then convert every question into a question vector $V = [v_{1}, \;  v_{2} \ldots \; v_{r}]$, where $v_{i} \in \{1, \; 0\}$ indicates the presence or absence of the $i^{th}$ question identifier word in the question. 
        
            \item \textbf{Tf-idf Vector:} We use tf-idf (term frequency - inverse document frequency) to assess how significant $w_{j}$ is to a $q_{i}$ in $Q$. It can be calculated as:
            \begin{eqnarray}
            \label{eq:tf-idf}
        	    tf \textrm{-} idf(w_{j},q_{i}) \; = \; \frac{f(w_{j},q_{i})}{\sum_{j = 1}^{m} f(w_{j},q_{i})} \; * \; log_{e}\frac{n}{\sum_{i = 1}^{n} f(w_{j},q_{i})}
            \end{eqnarray}
            where $f(w_{j},q_{i})$ is the frequency of $w_{j}$ in $q_{i}$, $n$ is the number of words in question $q_i$. 
            From the entire vocabulary, we consider only top $m^{'}$ words with the highest tf-idf values. We then convert every $q_{i}$ in the list Q into tf-idf vector such that position of every $w_{j}$ in $q_{i}$ is represented by $tf \textrm{-} idf(w_{j},q_{i})$.
        \end{itemize}
    We concatenate both the feature vectors to represent a question.
    \paragraph{{\bf Question Classifier:}} 
    \label{sec:question-classifier}
     The  SVM~\citep{cortes1995support,cristianini2000introduction} is a statistical classification technique and inspired by it's performance in~\citep{wang2018support,zhi2018support}. we use SVM learner as the base classifier. It takes the question feature vectors as input during the training stage to segregate the questions according to its type. It is a linear function which can be represented as,
        \begin{eqnarray}
        \label{eq:svm}
            f(v_{i}) \; = \; \langle v_{i} , w^{T} \rangle + b, \;\;\;\;\; where \;\; \langle v_{i} , w^{T} \rangle \; = \; \lvert\lvert v_{i} \rvert\rvert \; \lvert\lvert w^{T} \rvert\rvert \; cos(\theta)
        \end{eqnarray} 
        where $v_{i}$, $w^{T}$, and $b$ are feature vector of $i^{th}$ question, weight, and bias respectively.  So, $\forall i; \;i \in [1, \; n]$, either of the following equations can be true based on $q\_type$.
        \begin{eqnarray}
        \label{eq:tf-idf1}
            \langle v_{i} , w^{T} \rangle + b \;\; \geq 1, \;\;\;\;\;  if (q\_type == Yes/No)
        \end{eqnarray} 
        or
        \begin{eqnarray}
        \label{eq:tf-idf2}
            \langle v_{i} , w^{T} \rangle + b \;\; \leq -1, \;\;\;\;\; if( q\_type == Others)
        \end{eqnarray} 
        
We use the SVM with linear kernel, and use `\textit{hinge}' as the loss function. The hinge loss $\ell(y)$ of the prediction $y=f(v_i)$ for a true $q\_type$ $t$ and a classifier score $y$ is defined as,
    \begin{eqnarray}
        \label{eq:hinge-loss}
        \mathcal{\ell}(y) = max(0,1-t*f(v_i))
    \end{eqnarray} 
    
    \subsubsection{Answer Prediction}
    \label{sec:Answer-Prediction}
   
        Answer prediction component is at the second level of our hierarchy, where two separate models $M = \{m_{1}, \; m_{2}\}$ deal with the problem of answer generation at the lowest level nodes, i.e. leaf nodes. While $m_{1} = Yes/No$ deals with the problem of producing simple \textit{Yes/No} answers from simple incoming queries at the first leaf node, $m_{2} = Others$, on the other hand, deals with complex queries to produce other expertise answers at the second leaf node. We extract the question and image feature vectors which are then fused together. Based on  $q\_type$ this model passes the fused vector through several layers to finally generate the answer. We outline these tasks with more details in the following subsections.

    \paragraph{{\bf Question Feature Extraction:}}
    \label{sec:question-feat-extraction}
   
\begin{figure}[!ht]
            \includegraphics[width=1.0\textwidth,height=0.28\textwidth]{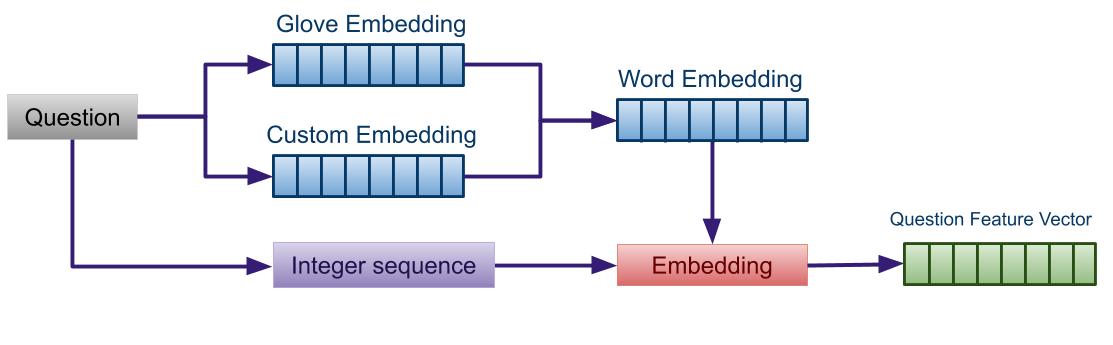}
        
            \caption{Flowchart of generating question embedding. The word embedding is the concatenation of GloVe and Custom (sub-word) embedding, which are used along with the integer sequence representation of the questions by the embedding layer.}
            \label{fig:question-embedding}
        \end{figure}

        We first pre-process the questions by converting the words into lowercase and then lemmatizing them to reduce the ambiguity among their different forms. Next we remove words like `\textit{the}', `\textit{and}', `\textit{with}' etc.  to discard useless information. We then map pure numbers to `\textit{num}' token and alphanumeric words to `\textit{pos}' token to minimize the complexity of information in questions. We then generate the integer sequence from the pre-processed questions that are finally fed to the embedding layer together with the word embedding for the extraction of question feature ($F^{Q}$) of dimension $m \times d$ where $m$ is the total number of words in the question and $d$ denotes the dimension of the word embedding vector. Fig~\ref{fig:question-embedding} illustrates the entire process.

    The components of the question feature extraction process are described below:
    \begin{itemize}
        \item {\bf Word embedding:} We use word embedding to vectorize words to capture their meaning. For this we first generate  $d_{1}$ dimensional vectors $G = [g_{1}, \; g_{2} \ldots \; g_{d_1}] $ using the GloVe~\citep{pennington2017stanford} vector. {We also introduce the sub-word embedding to capture the embedding of unknown word in medical terminology. For sub-word embedding, we follow the \citep{bojanowski-etal-2017-enriching} work on FastText vector and generate the sub-word embedding vector of dimension $d_2$ as $C = [c_{1}, \; c_{2} \ldots \; c_{d_2}] $.}
         We next concatenate the embeddings to create the final $d$-dimensional word embedding vector $E = [g_{1}, \; g_{2} \ldots \; g_{d_1}, \; c_{1}, \; c_{2} \ldots \; c_{d_2}] $. 
    \end{itemize}

    \paragraph{{\bf Image Feature Extraction:}}
    \label{sec:image-feature-extraction}
   
\begin{figure}[!ht]
            \includegraphics[width=0.72\textwidth,height=0.18\textwidth]{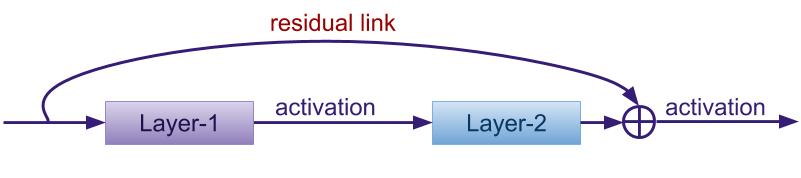}
            \caption{Canonical form of a 2-layer ResNet block. Layer-2 is skipped over activation from layer-1 using residual link.}
            \label{fig:residual-link}
        \end{figure}

        For image feature extraction, we use the Inception-Resnet-v2 model. It is a type of advanced CNN that integrates the inception module~\citep{Szegedy2014GoingDW} with ResNet. Inception enables one to accomplish a very good performance at comparatively low computational costs, while residual connections considerably speed up network training by enabling connection shortcuts. Together they allow the development of deeper and wider networks in the inception-resnet-v2 model. Basically, the network utilizes residual links~\citep{he2016deep} (Fig~\ref{fig:residual-link}) to combine filters of varying dimensions, which not only prevent the issue of degradation caused by deep structures but also decreases the training cost.

        Since the model expects an input of dimension $299\times299$, we re-size the input images to suit that dimension. We then initialize the model with weights pre-trained on ImageNet~\citep{russakovsky2015imagenet}. Such initialization facilitates transfer learning~\citep{long2017deep}, which is incorporated to enable a model to learn from another model pre-trained on a bigger dataset. It helps to train our deep neural network with a comparatively smaller dataset. Though the type of images in the medical domain is very different from those in the general domain, still transferring learned knowledge is more promising than training straight from scratch~\citep{tajbakhsh2016convolutional}. 
        Activations are obtained from the model's last fully connected layer as it represents the detected features of the medical image ($F^{I}$). The features generated are of size $1000$.

    \paragraph{{\bf Question and Image Feature Fusion:}}
    \label{sec:feature-fusion}
   
\begin{figure}[!ht]
            \includegraphics[width=0.8\textwidth,height=0.4\textwidth]{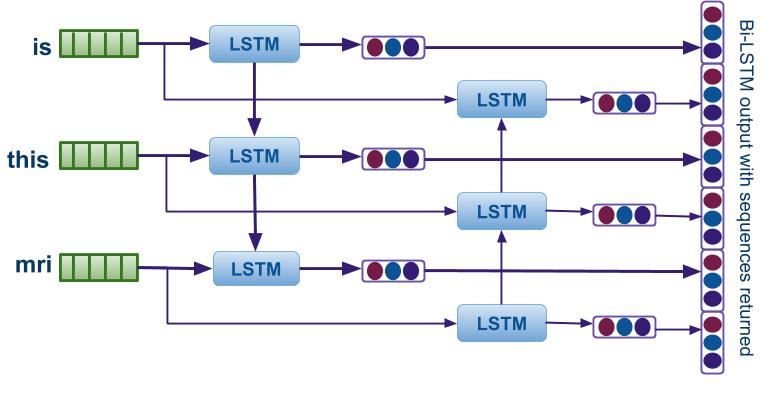}
    
            \caption{The processing of a question by Bi-LSTM to get the question representation. The word embedding of each word in the question is fed to a Bi-LSTM network, and the forward and backward hidden state outputs are concatenated at each time-step to get the final representation of the question. }
            \label{fig:bilstm}
        \end{figure}

        Unidirectional LSTM layer helps to capture the sequence information in the question, but it can only retain prior information as it has only seen the past inputs. In bidirectional layer inputs run bidirectionally in two ways, one from the past to the future and vice-versa. Therefore, before bi-modal feature fusion, we feed the extracted question feature vector $F^{Q}$ to the bidirectional layer with LSTM as input for the recurrent instance. The process to generate the representation of question by Bi-LSTM is depicted in Fig~\ref{fig:bilstm}.

        It helps to preserve the information from both past and future. We use it with sequences returned, so that LSTM hidden layer returns a sequence of values one per time-step instead of returning a single value for the entire sequence, such that $\overrightarrow{H} = [ \overrightarrow{h}_{1},\; \overrightarrow{h}_{2} \ldots \; \overrightarrow{h}_{m}]$, and  $\overleftarrow{H} = [ \overleftarrow{h}_{m},\; \overleftarrow{h}_{m-1} \ldots \; \overleftarrow{h}_{1}]$. Here, for forward and backward directions, $\overrightarrow{H}$, and  $\overleftarrow{H}$ are the sequence of hidden state outputs, while $\overrightarrow{h}_{i}$, and $\overleftarrow{h}_{i}$ are the hidden state outputs at $i^{th}$ time-step. The final Bi-LSTM output is then $\overleftrightarrow{H} = [ \overleftrightarrow{h}_{1},\; \overleftrightarrow{h}_{2} \ldots \; \overleftrightarrow{h}_{m}]$, with $\overleftrightarrow{h}_{i} = \overleftrightarrow{h}_{i} \odot \overleftrightarrow{h}_{m-i+1}; \; \forall i \in [1,m]$, where $\odot$ denotes the concatenate operator. To minimize the problem of overfitting due to small amount of training data we also used dropout value of 50\% in the this layer.

\begin{figure}[!ht]
            \includegraphics[width=0.9\textwidth,height=0.17\textwidth]{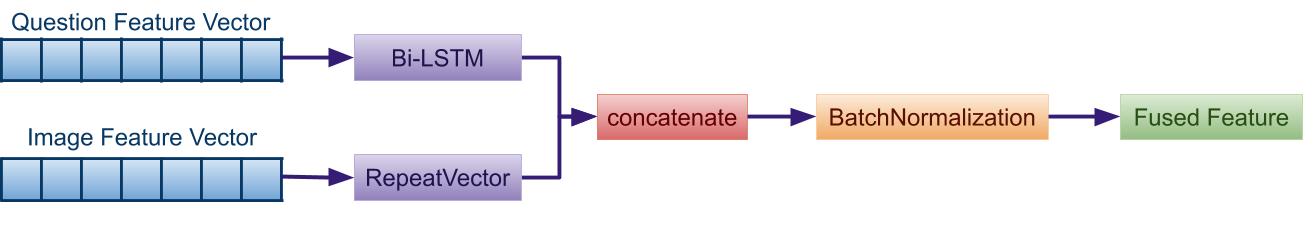}
            \caption{The architecture represents the fusion of the two modalities' feature vectors using concatenate layer, the output of which is normalized to generate the fused feature vector.}
            \label{fig:feature-fusion}
        \end{figure}

        We feed $F^{I}$ to the RepeatVector\footnote{RepeatVector(n) replicates the input feature vector n times.} layer to make it's dimension same as $F^{Q}$ for the modeling convenience. We then concatenate the repeated $F^{I}$, and $\overleftrightarrow{H}$ for fusion. We finally feed the output to Batch-Normalization\citep{Ioffe:2015:BNA:3045118.3045167} layer for regularization and to increase the stability of the network. Thus, the normalized fused feature ($F$) is:
        \begin{eqnarray}
        \label{eq:feature-fusion1}
            F = BN((Bidirectional(LSTM(F^{Q}))) \;\oplus\; (RepeatVector(m)(F^{I})))
        \end{eqnarray} 
        where, $BN$ and $\oplus$ represents Batch-Normalization and concatenation respectively. The process of feature fusion is illustrated in Fig~\ref{fig:feature-fusion}.

    \paragraph{\bf{Answer Prediction - Yes/No: }}
    \label{sec:answer-prediction-Yes/No}
   
         We treat model $m_{1}$ (c.f. Section~\ref{sec:Answer-Prediction}) as a two-class classification model. To generate the answer, we flatten the normalized fused feature ($F$) to generate a single long fused feature ($F^{'}$) which we pass through the fully-connected layer with two output neurons. We formulate the prediction procedure to predict the answer using Softmax as the activation function in the fully-connected layer as,
         
         \begin{equation}
          \label{eq:ans-Yes-No}
             \begin{aligned}
            \hat{a} &= \mathcal{P}(a_{i}|F^{'},\mathbf{W}, b) = softmax(F^{'}\mathbf{W}_{i}+b_{i})\\ &=\frac{e^{F^{'}\mathbf{W}_{i}+b_{i}}}{ e^{F^{'}\mathbf{W}_{Yes}+b_{Yes}}+e^{F^{'}\mathbf{W}_{No}+b_{No}}},\;\;i\in\{Yes,No\}
             \end{aligned}
         \end{equation}
        % \begin{eqnarray}
           
        % \end{eqnarray} 
        where, $\hat{a}$ is the prediction probability of selecting the $i^{th}$ answer word ($a_{i}$) given $F^{'}$ bias ($b_i$), and weight matrix $\mathbf{W}_i$ ($i\in\{Yes,No\}$). We use categorical cross entropy as loss function having the following formula.
        \begin{eqnarray}
            \label{eq:categorical-cross-entropy1}
            \mathcal{L}(a,\hat{a}) = - (a_{Yes}*\log(\hat{a}_{Yes})+a_{No}*\log(\hat{a}_{No}))
        \end{eqnarray} 
        where, $a_{i}$ and $\hat{a}_{i}$ denote the actual and predicted probability, respectively, of selecting $`Yes/No\textrm{'}$ as answer.

    \paragraph{{\bf Answer Prediction - Others:}}
    \label{sec:answer-prediction-others}
   
        We treat model $m_{2}$ (c.f. Section~\ref{sec:Answer-Prediction}) as a multi-label classification model for which we create a separate word-index dictionary for answers $D_a = \{w_{1} : 1, \;  w_{2} : 2 \ldots \; w_{z} : z\}$, where $z$ is the total number of unique words in the answer list. We also transform the $x^{th}$ answer list $A(x)$ of answer length $r$ to $A^{'}(x) = [a^{'}_{1}, \; a^{'}_{2} \ldots \; a^{'}_{r}]$, where $a^{'}_{i}$ is the $i^{th}$ answer which is encoded in the form of one-hot vector (a binary vector with values 0 and 1). We pass the normalized fused feature ($F'$) to the fully-connected layer with $t$ output neurons to generate $F^{''}$. We formulate the recursive prediction procedure to predict the answer words using TimeDistributed\footnote{\url{https://keras.io/api/layers/recurrent_layers/time_distributed/}} layer having Softmax activation as, 
        \begin{eqnarray}
            \label{eq:ans-others}
            \hat{a} = \mathcal{P}(a_{i}|F^{''},\hat{a}_{i-1},b) = softmax(F^{''}\mathbf{W}_{i}+b_{i}) = \frac{e^{F^{''}\mathbf{W}_{i}+b_{i}}}{\sum_{j = 1}^{z} e^{F^{''}\mathbf{W}_{j}+b_{j}}}
        \end{eqnarray} 
        where, $\hat{a}$ is the probability of selecting the $i^{th}$ answer word ($a_{i}$) given $F^{''}$, bias ($b$), and the set of probabilities of previously predicted answer words ($\hat{a}_{i-1}$), and $\mathbf{W}_i$ is the weight matrix. $z$ is the number of the words in vocabulary. We use categorical cross entropy as the loss function as follows: 
        \begin{eqnarray}
            \label{eq:categorical-cross-entropy}
            \mathcal{L}(a,\hat{a}) = -\sum_{i = 1}^{z}\sum_{j = 1}^{r} (a_{ij}*\log(\hat{a}_{ij}))
        \end{eqnarray} 
        In Eq~(\ref{eq:categorical-cross-entropy}), for $i^{th}$ word in the answer sequence of length $r$, $a_{ij}$ and $\hat{a}_{ij}$ denotes the actual and predicted probability of selecting the $j^{th}$ word of the answer dictionary having $z$ words.

    \subsubsection{Hyper-parameters}
    \label{sec:hyper-parameters}
   
        In the QS module to create the tf-idf vector, we consider top $500$ words with the highest tf-idf values from the vocabulary in the training set of $2000$ words. After studying the training set,  we select $10$ words (`\textit{is}', `\textit{was}', `\textit{are}', `\textit{how}', `\textit{can}', `\textit{does}', `\textit{which}', `\textit{what}', `\textit{type}', and `\textit{there}') to form the set of question identifier words. We use the default values of the rest of the parameters (e.g., the c in the SVM). For question embedding, we create $600$-dimensional word embedding by concatenating the two $300$-dimensional GloVe and FastText embeddings, following an approach similar to the work proposed in~\citep{ghannay2016word}. We create a question dictionary of size $1050$ to capture the most frequent words in the questions. For answers, we create a separate answer dictionary of size equal to the count of unique word in the answer list. As a negligible number of questions is of length greater than $21$, we fix the maximum question length as $21$. For answers having type `Others', we prune the maximum length to $11$. However, for \textit{Yes/No} type answers, the length is $1$, as only Yes or No is the probable answer. Consistency is maintained in the input length by appending `\textit{blank}' at the end for the shorter sequences and curbing longer sequences up to the required length. For the hidden layer of Bi-LSTM, we fix $128$ neurons in each direction. We use the Categorical Accuracy\footnote{https://github.com/keras-team/keras/blob/master/keras/metrics.py} as the metrics to calculate the mean accuracy rate for multiclass classification problems across all the predictions. We consider a batch of size $256$ for training. We set the number of epochs to $251$ and to optimize the weights during training we use Adam optimizer \citep{ADAMarticle}. We obtained the optimal hyper-parameters value based on the model performance on validation dataset.

    \subsubsection{Evaluation Metric}
    \label{sec:evaluation-metric}
   
        In our work, the datasets we use for evaluating our proposed model has only one reference answer. Thus, reporting high accuracy means generating a high number of the answers with exact same words as in the reference answer. This may not be necessary at all times and is a very complex task even in the medical domain. However, more than one answer may be correct. This is explained with an example given in Table~\ref{table:accuracy}.

        \begin{table}[!ht]
            \begin{tabular}{{  l  m{0.4\linewidth}  m{0.2\linewidth} }}
            
                {\bf Question} & {\bf Multiple correct answers} \\ \hline
                Where is the lung lesion located?
                & 
                \begin{itemize}
                    \itemsep-0.7em 
                        \item Right lobe
                        \item Lower lobe
                        \item Right lower lobe
                        \vspace{-0.5cm}
                    \end{itemize} 
                    \\\hline
                \end{tabular}
            \caption{Example where the absence of the degree of specification makes all the answers to the question are correct.}\label{table:accuracy}
        \end{table}
In this regard, we follow the standard evaluation schemes from the ImageCLEF2018 VQA-Med 2018 challenge. The evaluation metrics are as follows:
   \begin{itemize}
       \item \textbf{BiLingual Evaluation Understudy (BLEU) \citep{papineni2002bleu}:} It stands for  and it is a popular evaluation metric in machine translation, which compares the generated answer with the reference answer based on the number of n-grams of the generated answers that match with the reference answer, along with the brevity penalty for shorter output. BLEU is computed using the multiple modified n-gram provisions\footnote{To compute modified n-gram precision, all candidate n-gram counts and their
corresponding maximum reference counts are collected. The candidate counts are clipped by their corresponding reference maximum value, summed, and divided by the total number of candidate n-grams. Specifically,}
         \begin{equation}
             \text{BLEU} = \text{BP} \cdot \exp \bigg( \sum_{n=1}^{N} w_n \log_e p_n \bigg)
         \end{equation}
        where $p_n$ is the modified n-gram precision, BP is the brevity penalty to penalize short answer, $w_n$ is weight between $0$ and $1$ for $\log_e p_n$ and  $ \sum_{n=1}^{N} w_n=1$, $N$ is the maximum length of n-gram. BP can be computed as follows:
         \begin{equation}
 \text{BP} = 
\begin{cases} 
    1 & \text{if } c > r \\
    \exp \big(1-\frac{r}{c}\big) & \text{if } c \leq r
\end{cases}
\end{equation}
where $c$ is the number of unigrams in all the candidate answers $r$ is the best match lengths for each candidate answer in the dataset.

BLUE score serves as a better evaluation metric in this work, but it is not that effective when more than one medical term indicates the same part or symptom (e.g. the words `\textit{Lung}', and `\textit{Lobe}' refers to the same organ). As shown in Table~\ref{table:bleu}, all the answers are correct for the question, but the BLEU score decreases as it fails to consider synonyms during evaluation. 

        \begin{table}[!ht]
            \begin{tabular}{{  l  m{0.4\linewidth}  m{0.2\linewidth} }}
                {\bf Question} & {\bf Multiple correct answers} \\ \hline
                Where is the lesion located?
                & 
                 \begin{itemize}
                    \itemsep-0.7em 
                    \item Right lobe
                    \item Right lung
                    \item Right lobe of the lung
                    \vspace{-0.5cm}
                \end{itemize} 
                 \\ \hline
            \end{tabular}
            \caption{Example of the semantically similar answers. Although not all the answers to the question are the same, but they are semantically correct. }\label{table:bleu}
        \end{table}
\item {Word-based Semantic Similarity (WBBS) \citep{hasan2018overview}:} { It is another evaluation metrics used to assess the performance of the systems submitted in the VQA-Med 2018 challenge. WBSS metric based on Wu-Palmer Similarity\footnote{\url{https://datasets.d2.mpi-inf.mpg.de/mateusz14visualturing/calculate}} \cite{wu-palmer-1994-verb} with WordNet ontology in the backend. It computes a similarity score between the ground truth answer and system-generated answer by considering the word-level semantic similarity.}
  \end{itemize}
{We follow the evaluation setup discussed in ImageCLEF2018 VQA-Med 2018 challenge overview paper \citep{hasan2018overview} to evaluate the performance of the system. 
Towards this, we first pre-process the predicted and ground-truth answers and then calculate the scores. For pre-processing, we convert the answers to lower-case, remove the punctuations, and apply tokenization\footnote{\url{http://www.nltk.org/modules/nltk/tokenize/punkt.html\#PunktLanguageVars.word tokenize}} to the individual words in the answer. We also remove the stopwords in the answer from the NLTK's\footnote{\url{http://nltk.org/}} English stopword list. 
}
         
We also use the following metrics to  evaluate the effectiveness of our Answer Prediction module for $q\_type=`Yes/No\textrm{'}$ (c.f. Section~\ref{sec:answer-prediction-Yes/No}):
        
        \begin{itemize}
            \item {\bf Precision ($\mathrm{P}$):} It reflects the fraction of correctly predicted instances of a class (say $c$) from the total number of predicted instances as $c$.
            \begin{eqnarray}
                \label{eq:precision}
                \mathrm{P} = \dfrac{\mathopen|\{instances \; of\; c\} \cap  \{predicted \; instances \; as \; c\}\mathclose|}{\mathopen|\{predicted \; instances \; as \; c\}\mathclose|}
            \end{eqnarray}
            We report the macro-averaged precision $\mathrm{P_m}$, where $\mathrm{C}$ is a set of all the possible classes,
            \begin{eqnarray}
                \label{eq:macro-avg-precision}
                \mathrm{P_m} = (\mathrm \sum_{c\in C} {P_c})/\|C\|
                %+\mathrm{P}(q\_type=`No\textrm{'}))/2
            \end{eqnarray}
        
             \item {\bf Recall ($\mathrm{R}$):} It reflects the fraction of correctly predicted instances from the total number of actual instances belonging to $c$.
            \begin{eqnarray}
                \label{eq:recall}
                \mathrm{R} = \dfrac{\mathopen|\{instances \; of\; c\} \cap  \{predicted \; instances \; as \; c\}\mathclose|}{\mathopen|\{instances \; of\; c\}\mathclose|}
            \end{eqnarray}
            We report the macro-averaged recall ($\mathrm{R_m}$) for the set of all classes $\mathrm{C}$ similar to Eq~(\ref{eq:macro-avg-precision}).
            
            \item {\bf F$1$-score ($\mathrm{F1}$):} It is a function of $\mathrm{P}$ and $\mathrm{R}$. 
            \begin{eqnarray}
                \label{eq:f1-score} \mathrm{F{1}}=2*((\mathrm{P}*\mathrm{R})/(\mathrm{P}+\mathrm{R}))
            \end{eqnarray} 
            We report the macro-averaged F$1$-score   similar to Eq~(\ref{eq:macro-avg-precision}).
            
            \item {\bf Accuracy ($\mathrm{A}$):} It reflects the fraction of correctly predicted instances from the total number of instances.
            \begin{eqnarray}
                \label{eq:accuracy}
                \mathrm{A} = \dfrac{\mathopen|\{correctly \; predicted\; answers\}\mathclose|}{ \mathopen|\{answers\}\mathclose|}
            \end{eqnarray} 
        \end{itemize} 

\section{Data Description}
\label{sec:data-description}
   
        Datasets for Medical-VQA consists of Natural Language questions about the content of radiography images, and the task is to generate the appropriate answer. The questions are framed on the different modalities of medical image like \textit{`angiogram'}, \textit{`magnetic resonance imaging'}, \textit{`computed tomography'}, \textit{`ultrasound'}, etc. that describes how the image is taken. These images can have different orientations e.g. \textit{`sagittal'}, \textit{`axial'}, \textit{`longitudinal'}, \textit{`coronal'}, etc. Along with the variety in orientation and modalities, images can be of any body part or organ such as \textit{heart}, \textit{lung}, \textit{skull}, etc. (Fig~\ref{fig:sample-images}). 

         \begin{figure}[!ht]
     \captionsetup[subfigure]{labelformat=empty}
                \begin{subfigure}{.115\textwidth}
                    \includegraphics[width=\textwidth,height=\textwidth]{images/sample-imgs/brain.jpg}
                    \caption{(a) Brain}
                \end{subfigure}
                \begin{subfigure}{.115\textwidth}
                    \includegraphics[width=\textwidth,height=\textwidth]{images/sample-imgs/breast.jpg}
                    \caption{(b)Breasts}
                \end{subfigure}
                \begin{subfigure}{.115\textwidth}
                    \includegraphics[width=\textwidth,height=\textwidth]{images/sample-imgs/chest.jpg}
                    \caption{(c) Chest}
                \end{subfigure}
                \begin{subfigure}{.115\textwidth}
                    \includegraphics[width=\textwidth,height=\textwidth]{images/sample-imgs/CNS.jpg}
                    \caption{(d) CNS}
                \end{subfigure}   
                \begin{subfigure}{.115\textwidth}
                    \includegraphics[width=\textwidth,height=\textwidth]{images/sample-imgs/ap.jpg}
                    \caption{(e) AP}
                \end{subfigure}    
                \begin{subfigure}{.115\textwidth}
                    \includegraphics[width=\textwidth,height=\textwidth]{images/sample-imgs/axial.jpg}
                    \caption{(f) Axial}
               \end{subfigure}    
               \begin{subfigure}{.115\textwidth}
                    \includegraphics[width=\textwidth,height=\textwidth]{images/sample-imgs/coronal.jpg}
                    \caption{(g)Coronal}
               \end{subfigure}    
               \begin{subfigure}{.115\textwidth}
                    \includegraphics[width=\textwidth,height=\textwidth]{images/sample-imgs/pa.jpg}
                    \caption{(h) PA}
               \end{subfigure}
               \smallskip
                \caption{Sample images in the Medical-VQA dataset. The images in this dataset can be of different organs (a to d) and/or modalities (e to h).}
                \label{fig:sample-images}
            \end{figure}

        % For the proposed problem, we use RAD and CLEF18 datasets.
    \subsection{RAD Dataset}
    \label{sec:RAD-dataset}
   
        RAD dataset is a recently released dataset for VQA in the medical domain. Statistics of the dataset are as follows:
        \begin{itemize}
            \itemsep-0.5em 
            \item The training set consists of $3,064$ question-answer pairs.
            \item The test set consists of $451$ question-answer pairs.
        \end{itemize}
        Some of the images in the dataset are blurred while others contain markings such as short information, tags, etc. But none of the images in the dataset contains a stack of sub-images. The questions are primarily categorized into $11$ categories viz. abnormality, attribute, color, counting, modality,  organ system, other, plane, positional reasoning, and size. The average question length is $5$ to $7$ words which is greater than the answer length. $53\%$ of the answers are of \textit{Yes/No} type while rest of them are either one word or short phrase answers.

\begin{figure}[!ht]
            \begin{subfigure}{.5\textwidth}
                \includegraphics[width=1.0\textwidth,height=0.75\textwidth]{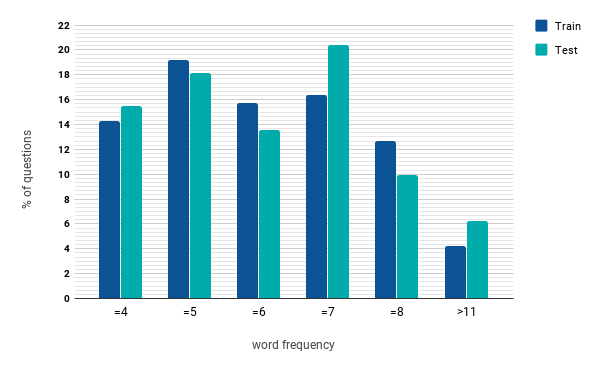}
                \caption{Questions}
                \label{fig:questions-RAD1}
           \end{subfigure}
           \begin{subfigure}{.5\textwidth}
                \includegraphics[width=1.0\textwidth,height=0.75\textwidth]{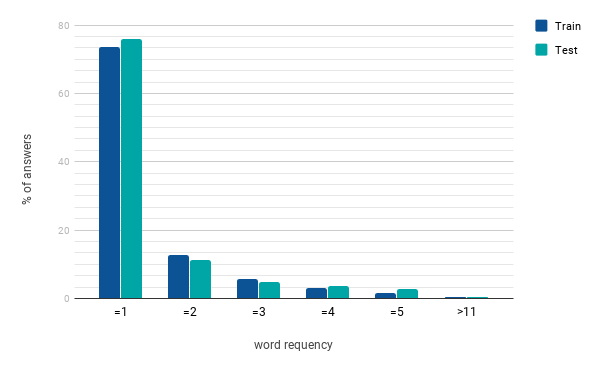}
                \caption{Answers}
                \label{fig:answers-RAD}
           \end{subfigure}
            \caption{ Word-Frequency distribution in the RAD dataset. The graph demonstrates that for both the questions and the answers, this distribution is almost similar for train and test data splits.}
            \label{fig:word-frequency-distribution-RAD}
        \end{figure}

        The maximum question length in the dataset is about $21$ words, with an average of $7$ words. It should also be noted that many questions are being rephrased, which are semantically similar. For example,
        \begin{itemize}
            \itemsep-0.5em 
            \item What is the size and density of the lesion?
            \item Describe the size and density of this lesion?
        \end{itemize}
        From the statistical study of the dataset, we find that only  $87\%$ of the free-form, and $93\%$ of the rephrased questions are unique, while only $32\%$ answers are unique. More than half of the answers are of \textit{Yes/No} type.
        This is visualized by the peak in the Fig~\ref{fig:answers-RAD} for one-word answers.
        
    \subsection{CLEF18 Dataset}
    \label{sec:CLEF18-dataset}
   
        CLEF18 task is similar to RAD task where a semi-automatic approach is used to generate the questions and answers from the captions. It uses the radiology images and their respective captions extracted from the PubMed Central articles\footnote{\url{https://www.ncbi.nlm.nih.gov/pmc/}} (essentially a subset of the ImageCLEF2017 caption prediction task \citep{eickhoff2017overview}). Due to the way the question-answer (QA) pairs are generated, they are diverse and descriptive. The dataset also contains a lot of artificial questions that are semantically invalid. Table~\ref{table:sample-training-data-CLEF18} demonstrates the complexity of a sample question-answer pair from the training data.

        \begin{table}[!ht]
        \begin{adjustwidth}{-0.9in}{-1.2in}
            \centering
            \begin{tabular}{{ m{0.465\linewidth}  m{0.465\linewidth} }}
                {\bf Question} & {\bf Answer} \\ \hline
                what reveals prominent bilateral enhancing parietal occipital lesions on flair and t2 sequences and small areas of hyperintensity in the left periventricular white matter on diffusion weighted images?
                & 
                mri of the brain
                \\ \hline
                what does mri in sagital plane show?
                & 
                the collection was superficial to the muscles of the back and the gluteal region but deep to the posterior layer of the thoraco lumbar fascia
                \\ \hline
            \end{tabular}
            \caption{Sample examples from the CLEF18 training data. Most of the questions and answers are descriptive and complicated in this dataset as they are generated semi-automatically. }\label{table:sample-training-data-CLEF18}
            \end{adjustwidth}
        \end{table}
         Statistics of the provided dataset are as follows:
        \begin{itemize}
            \item The training set consists of $5413$ questions along with their respective answers about $2,278$ images.
            \item The validation set consists of $500$ questions along with their respective answers about $324$ images.
            \item The test set consists of $500$ questions about $264$ images.
        \end{itemize}

\begin{figure}[!ht]
            \begin{subfigure}{.32\textwidth}
                \includegraphics[width=0.95\textwidth,height=0.75\textwidth]{images/image-type/blurred.jpg}
                \caption{Blurred}
                \label{fig:blurred-img}
           \end{subfigure}
           \begin{subfigure}{.32\textwidth}
                \includegraphics[width=0.95\textwidth,height=0.75\textwidth]{images/image-type/marking.jpg}
                \caption{Radiology markings}
                \label{fig:markings}
           \end{subfigure}
           \begin{subfigure}{.32\textwidth}
                \includegraphics[width=0.95\textwidth,height=0.75\textwidth]{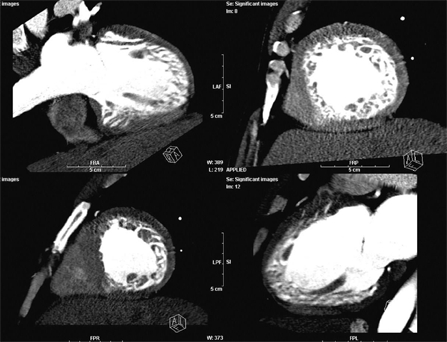}
                \caption{Stacked}
                \label{fig:stacked}
           \end{subfigure}
            \caption{Sample images in the CLEF18 dataset. Some of the images in this dataset are blurred (hazy/not clear), and/or contains short information in the form of radiology markings, and/or contains stack of sub-images.}
            \label{fig:sample-dataset-images}
        \end{figure}

        Some of the images in the dataset are blurred (Fig~\ref{fig:blurred-img}) and most of the images contain radiology markings (Fig~\ref{fig:markings}) such as short information, tags, arrows, etc. A few of them even consists of a stack of sub-images (Fig~\ref{fig:stacked}).
 
\begin{figure}[!ht]
        \begin{subfigure}{.5\textwidth}
            \includegraphics[width=1.0\textwidth,height=0.75\textwidth]{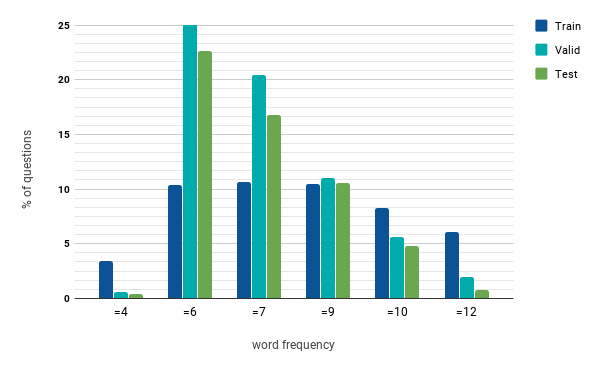}
            \caption{Questions}
            \label{fig:questions-CLEF18}
       \end{subfigure}
       \begin{subfigure}{.5\textwidth}
            \includegraphics[width=1.0\textwidth,height=0.75\textwidth]{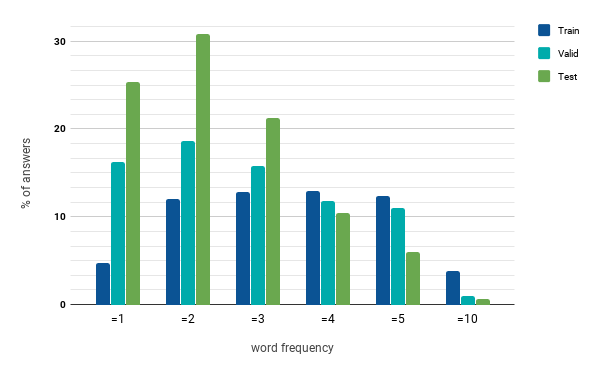}
            \caption{Answers}
            \label{fig:answers-CLEF18}
       \end{subfigure}
        \caption{Word-Frequency distribution in the CLEF18 dataset. The graph shows that for the data splits, this distribution is not comparable for both the questions and the answers.}
        \label{fig:word-frequency-distribution-CLEF18}
    \end{figure}

    Question categorization is not present in the dataset and only $0.6\%$ of the answers in training, $6\%$ in validation, and $10\%$ in test data are of \textit{Yes/No} type. From Fig~\ref{fig:word-frequency-distribution-CLEF18}, we can see that the average question length is more than answers, and the word frequency distribution is not even in the data splits.  

    \section{Results and Analysis}
    \label{sec:results-and-analysis}
    \subsection{Baselines}
    \label{sec:baselines}
{We compare our proposed methodology with the existing works in VQA-Med. Towards this, we use the following baseline models.}
        \begin{enumerate}

            \item {ResNet152 +  LSTM +  MFH \citep{peng2018umass}:} {We compare our approach with the best system reported in the ImageCLEF2018 challenge of VQA-Med 2018. They used LSTM to extract the question features, whereas the image features were extracted from the ResNet152 model pre-trained on the Imagenet dataset. For question and image feature fusion, they employed the co-attention mechanism with MFH to generate the question-image representation. They predicted the probable words to form the answers using the multi-label classification and then generated the answers using sampling.}
            
            \item {Inception-Resnet-v2 + Bi-LSTM +  Attention \citep{zhou2018employing}:} {Our second baseline model corresponds to the second best system participated in the ImageCLEF2018 challenge of VQA-Med 2018. Before applying the question and image to their proposed model, they performed the pre-processing steps on image and question both. They employed  Inception-Resnet-v2 model to extract image features, and Bi-LSTM model to encode the questions. They utilized the attention mechanism to fuse the image and question features.
}
            \item {VGG-16 + LSTM + SAN \citep{abacha2018nlm}:} {This is the third best system participated in the ImageCLEF2018 challenge of VQA-Med 2018.
            They used LSTM to extract the question features, whereas the image features were extracted from the last pooling layer of VGG-16 pre-trained on the Imagenet dataset. The stacked attention network \citep{yang2016stacked} is used to fuse the question and image features to obtain a single feature representation. These features are used to predict the answers from the given answer list, which is compiled from the training dataset.}
            
       \item {VGG16 + BERT + MFB \citep{yan2019zhejiang}:} {We compare the performance of our proposed model with the  best performing system participated in the ImageCLEF2019 challenge of VQA-Med 2019. They utilized the BERT model to extract the question features, whereas the image features were extracted from the multiple pooling layers of VGG-16 pre-trained on Imagenet dataset. Multi-Modal Factorized Bilinear (MFB) \citep{Yu_2017_ICCV} pooling were used to fuse the question and image features.  }
        \end{enumerate}
\begin{table}[h]
 \centering
\resizebox{\textwidth}{!}{%
    
    \begin{tabular}{l c c c|c c c|c c c}
    
        {}& \multicolumn{3}{c|}{\bf RAD}& \multicolumn{3}{c|}{\bf CLEF18}& \multicolumn{3}{c}{\bf CLEF18+RAD}\\
        \cline{2-10}
        
        {}& Yes/No & Others & Overall & Yes/No & Others & Overall& Yes/No & Others & Overall\\
        \thickhline
        
        \bf{Precision}& 0.98  &  0.99  &  0.99 & 1.00  &  0.93  &  0.93& 0.99  &  1.00  &  0.99\\
                \hline
                
        \bf{Recall}& 0.99  &  0.98  &  0.99& 0.28  &  1.00  &  0.93& 0.99  &  1.00  &  0.99\\
        \hline
        
        \bf{F\textsubscript{1}-score}& 0.99  &  0.98  &  0.99& 0.44 &  0.96  &  0.91&0.99  &  1.00  &  0.99\\
        \hline
    \end{tabular}
    }
    \caption{Performance of Question-Segregation model in terms of Precision, Recall, and F\textsubscript{1}-score.}
    \label{table:RADandRAD+CLEF18-QS}
\end{table}
\begin{table}[]
\centering
\resizebox{\textwidth}{!}{%
\begin{tabular}{l|cc|cc|cc}
% \hline
\multirow{2}{*}{\textbf{Model}} & \multicolumn{2}{c|}{\textbf{RAD}} & \multicolumn{2}{c|}{\textbf{CLEF18}} & \multicolumn{2}{c}{\textbf{CLEF18+RAD}} \\ \cline{2-7}
 & \textbf{BLEU} & \textbf{WBSS}  & \textbf{BLEU} & \textbf{WBSS}  & \textbf{BLEU} & \textbf{WBSS}  \\ \hline
\cite{peng2018umass} & -- & -- &  $0.161$ & $0.184$ &--&-- \\ 
\cite{peng2018umass}$^\mathbf{*}$ & $0.023$ & $0.104$ & $0.023$ & $0.072$  & $0.027$& $0.081$ \\ 
\cite{zhou2018employing}  & -- & -- &$0.134$ &$0.173$&-- & --\\ 
\cite{zhou2018employing}$^\mathbf{\$}$ & $0.522$ & $0.532$ & $0.072$ & $0.112$ & $0.277$ & $0.299$\\ 
\cite{abacha2018nlm}  & -- & -- & $0.121$ &$0.174$&--&--\\ 
\cite{abacha2018nlm}$^\mathbf{*}$  & $0.035$ & $0.213$ &  $0.051$ & $0.170$& $0.036$  & $0.173$\\ 
\cite{yan2019zhejiang}$^\mathbf{*}$ & $0.002$ & $0.011$  & $0.005$ &$0.069$ & $0.002$ & $0.023$ \\ 
Ours  & $0.411$ & $0.437$ & $0.132$ & $0.162$ & $0.257$ & $0.288$ \\ \hline \hline
\end{tabular}%
}
\caption{Comparison between the baseline models and our model in terms of BLEU and WBBS scores. Star ($^\mathbf{*}$) denote the re-implementation of the proposed work with the authors reported experimental setups. Dollar ($^\mathbf{\$}$) denote the official implementation of the approach proposed by the author.}
\label{table:RADandRAD+CLEF18}
\end{table}
\subsection{Result} \label{sec:result} Table~\ref{table:RADandRAD+CLEF18-QS} shows the performance of our QS model on RAD, CLEF18, and CLEF18+RAD dataset in terms of Precision, Recall, and F\textsubscript{1}-score. QS using SVM shows impressive results on the stated datasets. For CLEF18 dataset Recall and F\textsubscript{1}-score for `\textit{Yes/No}' type question is little less due to the less number of such questions in training example.

{Table~\ref{table:RADandRAD+CLEF18} shows the comparison of our proposed approach with the baseline models on the datasets in terms of BLEU and WBBS  scores. Our proposed approach for medical visual question answering achieves the BLEU score of $0.132$ and WBBS score of $0.162$ on the CLEF dataset. \citet{peng2018umass} reports the BLEU score of $0.161$ on the CLEF dataset. Since there is no official open-source implementation available for their system, we re-implemented the approach with the official experimental setup discussed in the paper, but only achieves the BLEU score of $0.023$ and WBBS score of $0.072$. We also fine-tune the hyper-parameters of their network with the available validation dataset, but we were not able to improve the performance further. \citet{peng2018umass} used the sampling approach to generate the answer in contrast with the other systems that participated in the ImageCLEF2018 VQA-Med 2018 challenge. The comparatively less (only $5413$ examples) amount of data to train the system raises the question of the effectiveness of the sampling approach to generate the correct answer. We also extend the experiments for the RAD and CLEF+RAD dataset on the re-implementation of the approach of \citet{peng2018umass}. For both the datasets, we randomly select $10\%$ of the data from the training set to fine-tune the network hyper-parameters. We obtain the BLEU score of $0.023$ and WBBS score of $0.104$ on the RAD dataset. The improvement (in terms of WBBS) in the RAD dataset can be understood by the fact that the RAD test set contains questions, which can be answered in a single-word. A similar observation is made on the results of the CLEF+RAD dataset.}

{\citet{zhou2018employing} reported the performance in terms of BLEU ($0.134$) and WBBS ($0.173$) scores on the CLEF dataset. We evaluated the performance of their implementations\footnote{\url{https://github.com/youngzhou97qz/CLEF2018-VQA-Med}} on the CLEF dataset, and recorded the BLEU and WBBS score of $0.072$ and $0.112$, respectively. We achieve the BLEU score of $0.522$ and WBBS score of $0.532$ on the RAD dataset with the re-implementation of their approaches. The reason for significant improvement for the RAD dataset is that \citet{zhou2018employing} performed considerable pre-processing on image, question, and answer and also post-processed the generated answers. In their pre-processing steps, they adopted image enhancement and reconstructed the images with exceedingly small random rotations, offsets, scaling, clipping, and increase
to $20$ images per image. For questions, they utilized the methods like stemming and lemmatization to alter verbs, nouns, and other words into their original forms. Furthermore, they replaced all the medical terms with their abbreviations, a combination of letters and numbers are replaced with `\textit{pos}' token and the pure numbers are
mapped to an `\textit{num}' token. For answers, they used lemmatization and removed all the stop words. They also replaced all the words associated with any number in answer. These words are generally the measures, for e.g.  `\textit{cm}' in `\textit{5 cm}'. As a post-processing step, they added several simple rules to the generated answers to make these more reasonable. In addition, they also deleted extra prepositions and additional words from the answers of yes/no questions. These additional pre-processing and rule-based post-processing steps make the system highly focused on medical VQA and not adaptable to the other domains of VQA. Additionally, the rules favor the short answers, which are only useful in the case of answer prediction and may suffer for answer generation. In contrast, our proposed approach achieves better performance for the CLEF dataset and comparable performance on the CLEF+RAD dataset without any additional processing on questions, images, or answers. Our system is generic, and it can be adaptable to any domain of VQA. }

{We also perform an additional experiment to further examine the performance improvement by \cite{zhou2018employing} on RAD and CLEF+RAD datasets. Toward this, we introduce an attention mechanism similar to the \cite{zhou2018employing} in our proposed model. The introduction of basic attention leads to significant performance improvement on the RAD  dataset. With the new model (our proposed + attention), we achieve $0.542$ and $0.553$ BLEU and WBBS scores, respectively, on the RAD dataset. The new model achieves the $0.110$ ($0.313$) and $0.051$ ($0.142$) BLEU (WBBS) scores on CLEF and CLEF+RAD datasets, respectively. The experiment with the new model shows that the attention favors the RAD dataset well, and we achieve much better BLEU, and WBBS scores compared to \cite{zhou2018employing}. However, it goes against the CLEF dataset, and we report the performance degradation on CLEF and CLEF+RAD datasets.}

{\citet{abacha2018nlm} reported the BLEU score of $0.121$ on the CLEF dataset. The official implementation of their system is not available. Therefore, we re-implemented the approach with the official experimental setup discussed in the paper, but only achieved the BLEU score of $0.051$ and WBBS score of $0.170$. Similar to the re-implementation of \citet{peng2018umass}, we fine-tune the hyper-parameters of their network with the available validation dataset.
In our re-implementation, we use the fine-tuned hyper-parameters to train the network on the training and validation dataset of CLEF (as done by \citet{abacha2018nlm}) to evaluate the performance on the test dataset of CLEF. 
We perform experiments with the RAD and CLEF+RAD datasets on the re-implemented approach of \citet{abacha2018nlm}. For both the RAD datasets, we randomly select $10\%$ of the data from the training set to fine-tune the network hyper-parameters. For CLEF+RAD dataset, we fine-tune the hyper-parameters on the combination of validation data of CLEF and $10\%$ of the RAD training set. We obtain the BLEU score of $0.0351$ and WBBS score of $0.213$ on the RAD dataset. For CLEF+RAD dataset, we report the BLEU and WBBS scores of $0.0365$ and $0.173$, respectively.
}

{We also compare the performance of our system with \cite{yan2019zhejiang}. Similar to the other works, we re-implement all the existing approaches and report the results on all the datasets. 
By observing performance on all the experiments, we conclude that the number of training samples is not sufficient enough to use for the sophisticated language-vision fusion mechanism such as SAN, MFB, or MFH. The model quickly gets over-fitted, which leads to lower performance on the test dataset. However, much simpler models (our proposed and \cite{zhou2018employing}) without any sophisticated language-vision fusion mechanism performs comparatively better on the task.}

{We follow the evaluation setup discussed in ImageCLEF2018 VQA-Med 2018 challenge overview paper \citep{hasan2018overview} to evaluate the performance of the system. But we cannot directly compare the results of our implementation of the ImageCLEF2018 VQA-Med 2018 participating systems \citep{peng2018umass,zhou2018employing,abacha2018nlm} and our proposed approach, as the participants did not use their own evaluation setup/script. The ImageCLEF2018 VQA-Med 2018 challenge organizers release the performance scores in terms of BLEU and WBBS scores. Further, the official evaluation scores are not readily available for use. %we tried to get the approached the organizers seeking for the official evaluation script. However, we did not hear back from them. 
To make fair comparison and reproducibility of our works, we make our source codes available\footnote{\url{https://bit.ly/3aH6EFm}} along with the evaluation script of all the re-implementations of the existing works and our proposed approach.
}

We also analyze that, in general, a model performs better if more training samples are present, thus the models must have better scores when trained and tested on the combined dataset. But further analysis of the results reveals that the models perform better on the individual datasets than on the combined one. This is due to the difference in the size of the datasets, and the way the questions are framed and answers are generated. While the CLEF18 dataset is created by a semi-automatic approach, the RAD dataset is manually created. The difference in quality and complexity of the generated sequence in the two datasets is clearly visible in Table~\ref{table:RAD,andCLEF+RAD}, where both the questions require liver identification, but there is a considerable difference in complexity of the question as well as the answer. Due to the complications and the limited number of examples in the datasets, the model fails to learn efficiently.

% \paragraph{\textbf{Comparison to the State-of-the-art works in Medical VQA}:}

        \begin{table}[!ht]
            \begin{tabular}{l  m{0.45\linewidth}  m{0.25\linewidth}}
                {} & {\bf Question}& {\bf Answer} \\ \hline
                \bf{RAD}
                &
                What solid organ is seen on the right side of this image?
                &
                The liver
                \\ \hline
                \bf{CLEF18}
                &
                What shows the dilated common bile duct with a filling defect within it indicating the tumor extending?
                &
                Magnetic resonance imaging image of the liver
                \\ \hline
            \end{tabular}
            
            \caption{Example of question-answer pair, which shows that question and answer structure in CLEF18 is more complicated than in RAD.}\label{table:RAD,andCLEF+RAD}
        \end{table}

    \subsubsection{Impact of QS module}
    \label{sec:impact-of-QS}
   
\begin{figure}[!ht]
            \begin{subfigure}{.32\textwidth}
                \includegraphics[width=1.0\textwidth,height=0.75\textwidth]{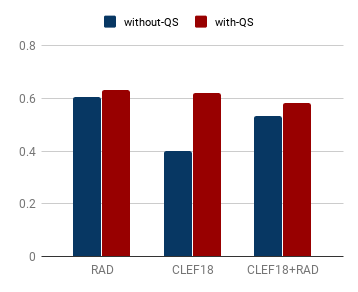}
                \caption{Yes/No}
                \label{fig:comp-yesno}
           \end{subfigure}
           \begin{subfigure}{.32\textwidth}
                \includegraphics[width=1.0\textwidth,height=0.75\textwidth]{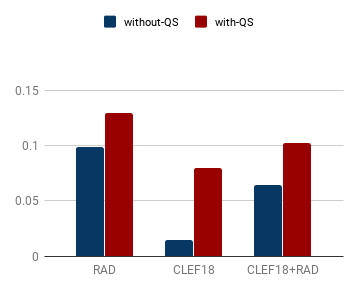}
                \caption{Others}
                \label{fig:comp-others}
           \end{subfigure}
           \begin{subfigure}{.32\textwidth}
                \includegraphics[width=1.0\textwidth,height=0.75\textwidth]{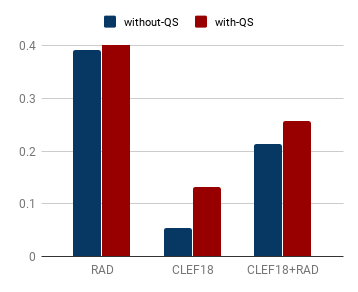}
                \caption{Overall}
                \label{fig:comp-overall}
           \end{subfigure}
            \caption{Impact of QS on the model performance. it shows that with QS the model performs better regardless of the type of question or dataset. }
            \label{fig:score-comp}
        \end{figure}

        Table~\ref{table:combined-results} demonstrates the impact of QS (c.f. Section~\ref{sec:question-segregation}) module. It also reveals that QS improves the model's performance by a significant margin regardless of the question type. The performance difference is clearly visible in Fig~\ref{fig:score-comp}. 

        \begin{table}[!ht]
            \begin{tabular}{l c c|c c|c c}
    
                {}& \multicolumn{2}{c|}{\bf RAD}& \multicolumn{2}{c|}{\bf CLEF18}& \multicolumn{2}{c}{\bf CLEF18+RAD}\\
                \cline{2-7}
        
                {}& without & with & without & with & without & with\\
                \thickhline
                
                \bf{Yes/No}& 0.606 & \bf{0.634} & 0.400& \bf{0.620} & 0.534 & \bf{0.581}\\
                \hline
                
                \bf{Others}& 0.099 & \bf{0.129} & 0.015& \bf{0.080} & 0.064 & \bf{0.102}\\
                \hline
                
                \bf{Overall}& 0.392 & \bf{0.411} & 0.053& \bf{0.132} & 0.213 & \bf{0.257}\\
                \hline
            \end{tabular}
    
            \caption{ Result of our model with and without question type segregation, for RAD, CLEF18, and CLEF18+RAD dataset.}
            \label{table:combined-results}
        \end{table}

    \paragraph*{{\bf Impact of QS on questions with type `\textit{Yes/No}':}}
    \label{sec:impact-of-QS-yes/no}
         For this type of question, the main advantage of QS is that it prevents the model from predicting any answer phrases other than a straightforward `\textit{Yes}' or `\textit{No}'. While a model without QS can predict any answer word or sequence of answer words that turns out to be irrelevant. Table~\ref{table:ans-pred-yes/no} shows the model's predicted responses with and without the QS module.

        \begin{table}[!ht]
            \begin{adjustwidth}{-0.9in}{-1.2in} 
            \centering
        
            \begin{tabular}{{m{0.15\linewidth}m{0.18\linewidth}m{0.18\linewidth}m{0.18\linewidth}m{0.18\linewidth}}}\\
            \thickhline
            \begin{minipage}{.5\textwidth}
              \vspace{0.02\linewidth}
              \vspace{0.02\linewidth}
            \end{minipage}
            & 
            \begin{minipage}{.5\textwidth}
              \vspace{0.02\linewidth}
              \includegraphics[width=0.5\linewidth, height=0.4\textwidth]{images/quality-que1/que1.jpg}
              \vspace{0.02\linewidth}
            \end{minipage}
            &
            \begin{minipage}{.5\textwidth}
              \vspace{0.02\linewidth}
              \includegraphics[width=0.5\linewidth, height=0.4\textwidth]{images/quality-que1/que2.jpg}
              \vspace{0.02\linewidth}
            \end{minipage}
            &
            \begin{minipage}{.5\textwidth}
              \vspace{0.02\linewidth}
              \includegraphics[width=0.5\linewidth, height=0.4\textwidth]{images/quality-que1/que3.jpg}
              \vspace{0.02\linewidth}
            \end{minipage}
            &
            \begin{minipage}{.5\textwidth}
              \vspace{0.02\linewidth}
              \includegraphics[width=0.5\linewidth, height=0.4\textwidth]{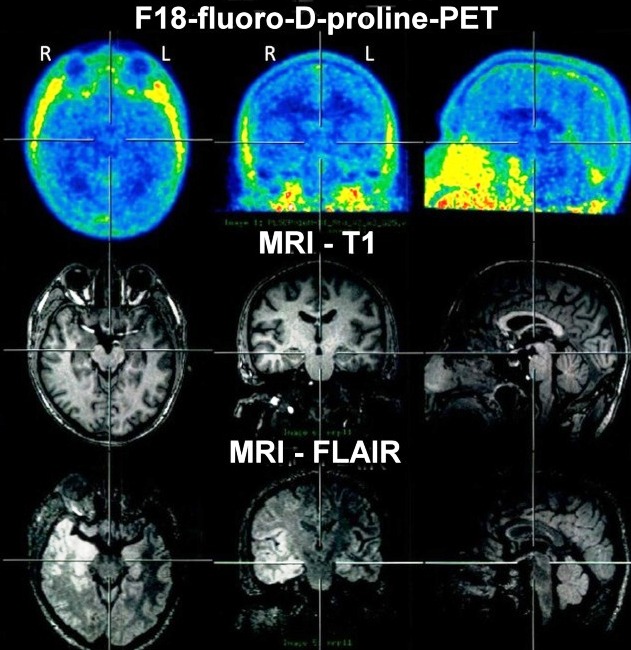}
              \vspace{0.02\linewidth}
            \end{minipage}
            \\ \hline
            \vspace{0.05\linewidth}
            {\bf Question\hspace{0.235\linewidth}:} & Is the GI tract is highlighted by contrast? & Is the surrounding phlegmon normal? & Were both sides affected? & Does the PET scan show abnormal tracer accumulation?\\ \hline
            \vspace{0.05\linewidth} 
            {\bf Ans (w/o.) :} & bilateral & bronchiectasis & axial & internal whorled\\
            \hline
            {\bf Ans (w.)\hspace{0.56cm}:} & Yes & Yes & Yes & No\\
            \hline
            {\bf Ans (GT)\hspace{0.15\linewidth}:} & Yes & No & Yes & No
            \\
            \thickhline
            \end{tabular}
            \caption{Comparison of Ground Truth (GT) answer with the predicted answer having question-type \textit{Yes/No} by the model with (w.) and without (w/o.) QS. Without QS, our model predicts answer words that are not expected for question in this category.  }
            \label{table:ans-pred-yes/no}
            \end{adjustwidth}
        \end{table}
        
        With QS, the search space is reduced to only two words that is `\textit{Yes}' and `\textit{No}' while a model without QS will have to unnecessarily predict the answer words from the entire answer dictionary (a dictionary containing all the possible answer words). This reduction in search space leads to a better chance of predicting the right answer. Table~\ref{table:precision-recall-f1} demonstrates the effectiveness of QS in our model for `\textit{Yes/No}' type questions. The scores show that, for RAD, CLEF18 and CLEF18+RAD datasets, QS improves precision (Eq.~\ref{eq:precision}), recall (Eq.~\ref{eq:recall}) and f1-score (Eq.~\ref{eq:f1-score}) by $0.3$, $0.1$, $0.2$ and $0.3$, $0.5$, $0.2$ points respectively.
         
        \begin{table}[!ht]
    \centering
        \begin{tabular}{l l l|l l|l l}
            {}& \multicolumn{2}{c|}{\bf RAD}& \multicolumn{2}{c|}{\bf CLEF18}& \multicolumn{2}{c}{\bf CLEF18+RAD}\\
            \cline{2-7}
            
            {}& w/o. & w. & w/o. & w. & w/o. & w.\\
            \thickhline
    
            \bf{precision}& 0.38 & \bf{0.63} & 0.60& \bf{0.72} & 0.37 & \bf{0.58}\\
            \hline
            
            \bf{Recall}& 0.37 & \bf{0.64} & 0.12& \bf{0.63} & 0.35 & \bf{0.58}\\
            \hline
                
            \bf{F1-score}& 0.37 & \bf{0.63} & 0.17& \bf{0.62} & 0.34 & \bf{0.58}\\
            \hline
        \end{tabular}
            \caption{ Performance of our model with (w.) and without (w/o.) QS module for Y/N type questions.}
            \label{table:precision-recall-f1}sample-imgs/chest
        \end{table}
 
    \paragraph*{{\bf Impact of QS on questions with type `\textit{Others}':}}
    \label{sec:impact-of-QS-others}        
        A model without QS can predict `\textit{Yes}' or `\textit{No}' as an answer which is completely irrelevant for the questions with type `\textit{Others}'. But the quality of prediction increases when QS is integrated with the same model. Table~\ref{table:ans-pred-others} includes several such instances where the model with QS fails to predict the answer correctly but produces an answer that is applicable to the question-image pair and is more acceptable than straightforward `\textit{Yes}' or `\textit{No}'.
        
        \begin{table}[!ht]
            \begin{adjustwidth}{-0.9in}{-1.2in} 
            \centering
            \begin{tabular}{{m{0.15\linewidth}m{0.18\linewidth}m{0.18\linewidth}m{0.18\linewidth}m{0.18\linewidth}}}\\
            \thickhline
            \begin{minipage}{.5\textwidth}
              \vspace{0.02\linewidth}
              \vspace{0.02\linewidth}
            \end{minipage}
            & 
            \begin{minipage}{.5\textwidth}
              \vspace{0.02\linewidth}
              \includegraphics[width=0.5\linewidth, height=0.4\textwidth]{images/quality-que2/que1.jpg}
              \vspace{0.02\linewidth}
            \end{minipage}
            &
            \begin{minipage}{.5\textwidth}
              \vspace{0.02\linewidth}
              \includegraphics[width=0.5\linewidth, height=0.4\textwidth]{images/quality-que2/que2.jpg}
              \vspace{0.02\linewidth}
            \end{minipage}
            &
            \begin{minipage}{.5\textwidth}
              \vspace{0.02\linewidth}
              \includegraphics[width=0.5\linewidth, height=0.4\textwidth]{images/quality-que2/que3.jpg}
              \vspace{0.02\linewidth}
            \end{minipage}
            &
            \begin{minipage}{.5\textwidth}
              \vspace{0.02\linewidth}
              \includegraphics[width=0.5\linewidth, height=0.4\textwidth]{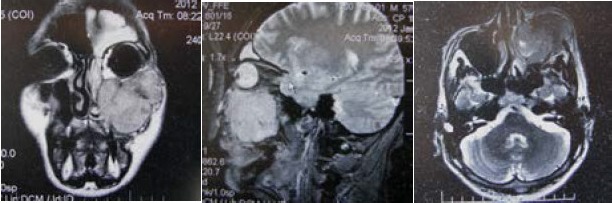}
              \vspace{0.02\linewidth}
            \end{minipage}
            \\ \hline
            \vspace{0.05\linewidth}
            {\bf Question\hspace{0.235\linewidth}:} & (1) The image is taken in what plane? & (2) Where are the infarcts? & (3) Is this patient male or female? & (4) What does the mri show?\\ \hline
            \vspace{0.05\linewidth}
            {\bf Ans (w/o.) :} & Yes & No & No & No in\\
            \hline
            {\bf Ans (w.)\hspace{0.54cm}:} & pa & in right & chest & mass\\
            \hline
            {\bf Ans (GT)\hspace{0.15\linewidth}:} & axial & basal ganglia & female & tumor
            \\ 
            \thickhline
            \end{tabular}
            \caption{ Comparison of Ground Truth (GT) answer (Ans) with the predicted answer having question-type \textit{Yes/No} by the model with (w.) and without (w/o.) QS. Without QS, our model predicts Yes or No along with other answer words that are not expected for question in this category. }
            \label{table:ans-pred-others}
            \end{adjustwidth}
        \end{table} 
        
                \begin{table}[]
            \begin{adjustwidth}{-0.9in}{-1.2in} 
            \centering
        
            \begin{tabular}{{m{0.15\linewidth}m{0.18\linewidth}m{0.18\linewidth}m{0.18\linewidth}m{0.18\linewidth}}}\\
            
            \thickhline
            \begin{minipage}{.5\textwidth}
              \vspace{0.02\linewidth}
              {\bf Semantic \\Error:}
              \vspace{0.02\linewidth}
            \end{minipage}
            & 
            \begin{minipage}{.5\textwidth}
              \vspace{0.02\linewidth}
              \includegraphics[width=0.5\linewidth, height=0.4\textwidth]{images/error-imgs/que11.jpg}
              \vspace{0.02\linewidth}
            \end{minipage}
            &
            \begin{minipage}{.5\textwidth}
              \vspace{0.02\linewidth}
              \includegraphics[width=0.5\linewidth, height=0.4\textwidth]{images/error-imgs/que12.jpg}
              \vspace{0.02\linewidth}
            \end{minipage}
            &
            \begin{minipage}{.5\textwidth}
              \vspace{0.02\linewidth}
              \includegraphics[width=0.5\linewidth, height=0.4\textwidth]{images/error-imgs/que13.jpg}
              \vspace{0.02\linewidth}
            \end{minipage}
            &
            \begin{minipage}{.5\textwidth}
              \vspace{0.02\linewidth}
              \includegraphics[width=0.5\linewidth, height=0.4\textwidth]{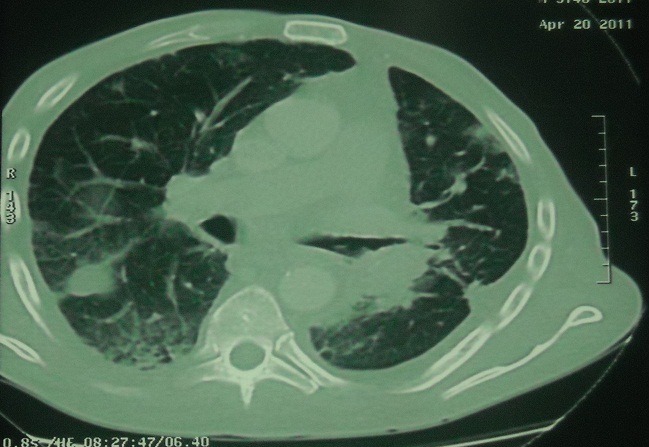}
              \vspace{0.02\linewidth}
            \end{minipage}
            \\ \hline
        
            \vspace{0.05\linewidth}
            {\bf Question\hspace{0.237\linewidth}:} & Where do you see acute infarcts? & How is the patient oriented? & How was this image taken? & where is the lesion located? \\ \hline
            \vspace{0.05\linewidth} 
            {\bf Ans(GT)\hspace{0.22\linewidth}:} & R frontal lobe & Posterior-Anterior & T2-MRI & Lower lobe of the right lung\\
            \hline
            {\bf Ans(pred.) :} & right frontal lobe & pa & mri t2 weighted & right lower lobe\\
            \hline
            {\bf Comments\hspace{0.21cm}:} & R refers to Right. & pa is acronym for Posterior-Anterior. & T2-MRI denotes mri (t2-weighted). & Semantically same answer with different sentence structure.\\
            \thickhline
            
            \begin{minipage}{.5\textwidth}
              \vspace{0.02\linewidth}
              {\bf Modality/\\Plane \\Confusion:}
              \vspace{0.02\linewidth}
            \end{minipage}
            & 
            \begin{minipage}{.5\textwidth}
              \vspace{0.02\linewidth}
              \includegraphics[width=0.5\linewidth, height=0.4\textwidth]{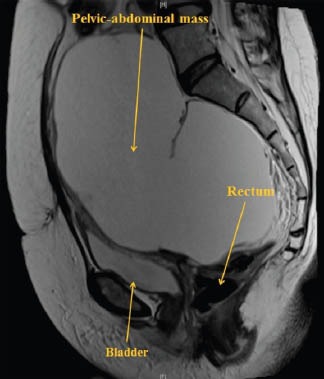}
              \vspace{0.02\linewidth}
            \end{minipage}
            &
            \begin{minipage}{.5\textwidth}
              \vspace{0.02\linewidth}
              \includegraphics[width=0.5\linewidth, height=0.4\textwidth]{images/error-imgs/que22.jpg}
              \vspace{0.02\linewidth}
            \end{minipage}
            &
            \begin{minipage}{.5\textwidth}
              \vspace{0.02\linewidth}
              \includegraphics[width=0.5\linewidth, height=0.4\textwidth]{images/error-imgs/que23.jpg}
              \vspace{0.02\linewidth}
            \end{minipage}
            &
            \begin{minipage}{.5\textwidth}
              \vspace{0.02\linewidth}
              \includegraphics[width=0.5\linewidth, height=0.4\textwidth]{images/error-imgs/que24.jpg}
              \vspace{0.02\linewidth}
            \end{minipage}
            \\ \hline
            
            \vspace{0.05\linewidth}
            {\bf Question\hspace{0.237\linewidth}:} & What imaging modality is this? & What kind of image is this? & What type of image is this? & What type of image is this?\\ \hline
            \vspace{0.05\linewidth} 
            {\bf Ans(GT)\hspace{0.22\linewidth}:} & Sagittal view of t2 weighted mri & X-ray & Plain film x-ray & CT with contrast\\
            \hline
            {\bf Ans(pred.) :} & mri t2 weighted & axial x ray & x ray & ct\\
            \hline
            {\bf Comments\hspace{0.21cm}:} & Identification of plane is not required as per the question. & Predicted answer is more precise. & GT answer is more precise. & GT answer consists of modality and it's subtype information.\\
            \thickhline
            
            \end{tabular}
            \caption{Error analysis (Semantic Error and Modality/Plane  confusion): Comparison of Ground Truth answer with the predicted answer for understanding the error types. }
            \label{table:error-analysis-1}
         \end{adjustwidth}
        \end{table}
        
        \begin{table}[]
            \begin{adjustwidth}{-0.9in}{-1.2in} 
            \centering
        
            \begin{tabular}{{m{0.15\linewidth}m{0.18\linewidth}m{0.18\linewidth}m{0.18\linewidth}m{0.18\linewidth}}}\\
            \thickhline
            
            \begin{minipage}{.5\textwidth}
              \vspace{0.02\linewidth}
              {\bf Specification \\Error:}
              \vspace{0.02\linewidth}
            \end{minipage}
            & 
            \begin{minipage}{.5\textwidth}
              \vspace{0.02\linewidth}
              \includegraphics[width=0.5\linewidth, height=0.4\textwidth]{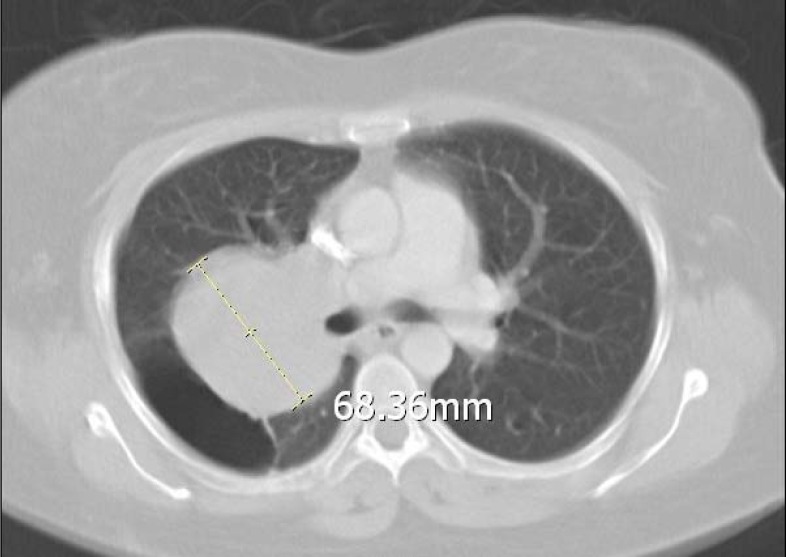}
              \vspace{0.02\linewidth}
            \end{minipage}
            &
            \begin{minipage}{.5\textwidth}
              \vspace{0.02\linewidth}
              \includegraphics[width=0.5\linewidth, height=0.4\textwidth]{images/error-imgs/que32.jpg}
              \vspace{0.02\linewidth}
            \end{minipage}
            &
            \begin{minipage}{.5\textwidth}
              \vspace{0.02\linewidth}
              \includegraphics[width=0.5\linewidth, height=0.4\textwidth]{images/error-imgs/que33.jpg}
              \vspace{0.02\linewidth}
            \end{minipage}
            &
            \begin{minipage}{.5\textwidth}
              \vspace{0.02\linewidth}
              \includegraphics[width=0.5\linewidth, height=0.4\textwidth]{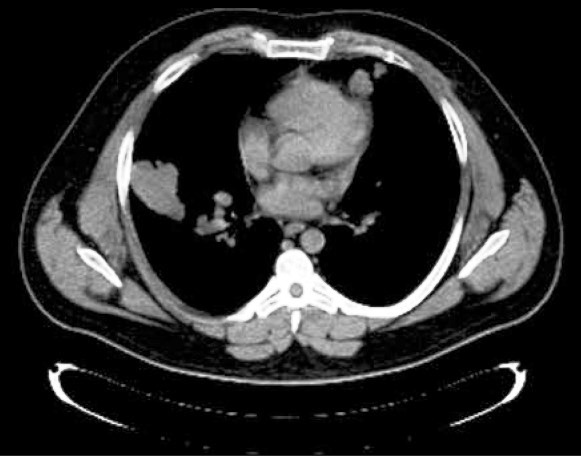}
              \vspace{0.02\linewidth}
            \end{minipage}
            \\ \hline
            
            \vspace{0.05\linewidth}
            {\bf Question\hspace{0.237\linewidth}:} & What does the ct scan of the chest show? & What lobe of the brain is the lesion located in? & What is the location of the lesion? & Where are the lesions formed?\\ \hline
            \vspace{0.05\linewidth} 
            {\bf Ans(GT)\hspace{0.22\linewidth}:} & A large mass & Right frontal lobe & Right lower lateral lung field & Mediastinum and Hilum of the right lung\\
            \hline
            {\bf Ans(pred.) :} & mass & right lobe & lower lung & in right lung\\
            \hline
            {\bf Comments\hspace{0.21cm}:} & Predicted answer does not specify the amount of mass. & Predicted answer lacks image-plan specification. & GT answer is more specific in terms of lesion location. & GT answer specifically defines the lesion formation.\\
            \thickhline
            \begin{minipage}{.5\textwidth}
              \vspace{0.02\linewidth}
              {\bf Boundary \\Loss:}
              \vspace{0.02\linewidth}
            \end{minipage}
            & 
            \begin{minipage}{.5\textwidth}
              \vspace{0.02\linewidth}
              \includegraphics[width=0.5\linewidth, height=0.4\textwidth]{images/error-imgs/que41.jpg}
              \vspace{0.02\linewidth}
            \end{minipage}
            &
            \begin{minipage}{.5\textwidth}
              \vspace{0.02\linewidth}
              \includegraphics[width=0.5\linewidth, height=0.4\textwidth]{images/error-imgs/que42.jpg}
              \vspace{0.02\linewidth}
            \end{minipage}
            &
            \begin{minipage}{.5\textwidth}
              \vspace{0.02\linewidth}
              \includegraphics[width=0.5\linewidth, height=0.4\textwidth]{images/error-imgs/que43.jpg}
              \vspace{0.02\linewidth}
            \end{minipage}
            &
            \begin{minipage}{.5\textwidth}
              \vspace{0.02\linewidth}
              \includegraphics[width=0.5\linewidth, height=0.4\textwidth]{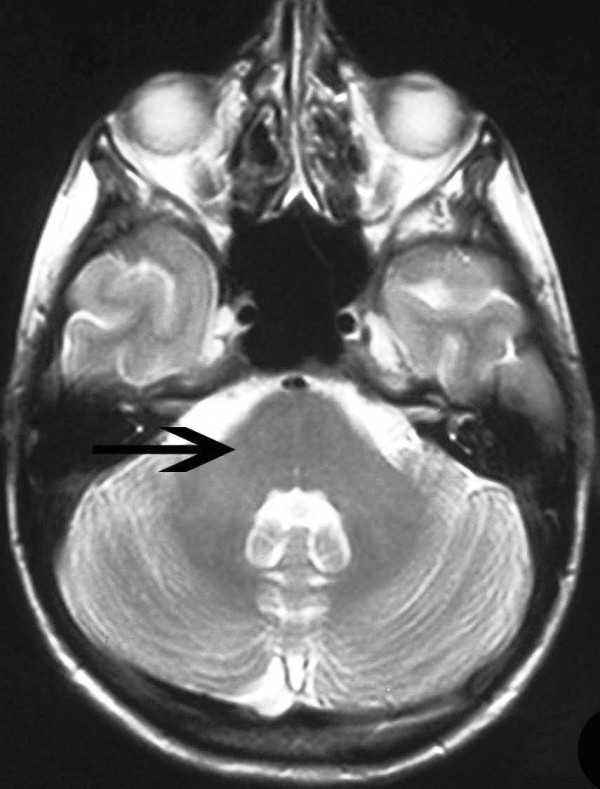}
              \vspace{0.02\linewidth}
            \end{minipage}
            \\ \hline
            
            \vspace{0.05\linewidth}
            {\bf Question\hspace{0.237\linewidth}:} & In what plane was this image taken? & What lobe of the brain is the lesion located in? & Is the spleen present? & What kind of image is this?\\ \hline
            \vspace{0.05\linewidth} 
            {\bf Ans(GT)\hspace{0.22\linewidth}:} & Axial plane & Right frontal lobe & On the patient's left & T2 weighted mri\\
            \hline
            {\bf Ans(pred.) :} & axial & right frontal & left & t2 weighted\\
            \hline
            {\bf Comments\hspace{0.21cm}:} & \textit{`plane'} is uninformative word in the GT answer. & Uninformative word \textit{`lobe'} in the GT answer. & GT answer consists additional and uninformative words. & Predicted answer lacks modality information.\\
            \thickhline
            
            \end{tabular}
            \caption{Error analysis (Specification Error and Boundary Loss): Comparison of Ground Truth answer with the predicted answer for understanding the error types.}
            \label{table:error-analysis-2}
         \end{adjustwidth}
        \end{table}
        
        \begin{table}[ht!]
            \begin{adjustwidth}{-0.9in}{-1.2in} 
            \centering
        
            \begin{tabular}{{m{0.15\linewidth}m{0.18\linewidth}m{0.18\linewidth}m{0.18\linewidth}m{0.18\linewidth}}}\\
            \thickhline
            \begin{minipage}{.5\textwidth}
              \vspace{0.02\linewidth}
              {\bf Miscellaneous \\Error:}
              \vspace{0.02\linewidth}
            \end{minipage}
            & 
            \begin{minipage}{.5\textwidth}
              \vspace{0.02\linewidth}
              \includegraphics[width=0.5\linewidth, height=0.4\textwidth]{images/error-imgs/que51.jpg}
              \vspace{0.02\linewidth}
            \end{minipage}
            &
            \begin{minipage}{.5\textwidth}
              \vspace{0.02\linewidth}
              \includegraphics[width=0.5\linewidth, height=0.4\textwidth]{images/error-imgs/que52.jpg}
              \vspace{0.02\linewidth}
            \end{minipage}
            &
            \begin{minipage}{.5\textwidth}
              \vspace{0.02\linewidth}
              \includegraphics[width=0.5\linewidth, height=0.4\textwidth]{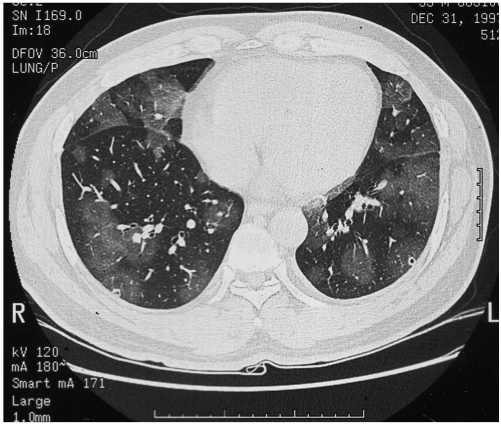}
              \vspace{0.02\linewidth}
            \end{minipage}
            &
            \begin{minipage}{.5\textwidth}
              \vspace{0.02\linewidth}
              \includegraphics[width=0.5\linewidth, height=0.4\textwidth]{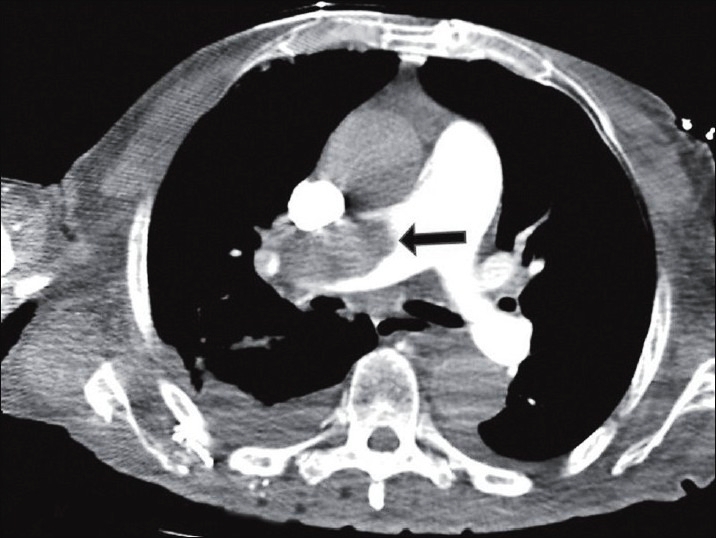}
              \vspace{0.02\linewidth}
            \end{minipage}
            \\ \hline
            
            \vspace{0.05\linewidth}
            {\bf Question\hspace{0.237\linewidth}:} & Is this an MRI or a CT scan? & Is this patient male or female? & Where are the lesions found? & What does the ct pulmonary angiogram show?\\ \hline
            \vspace{0.05\linewidth} 
            {\bf Ans(GT)\hspace{0.22\linewidth}:} & MRI & Female & In both lungs & Massive filling defect\\
            \hline
            {\bf Ans(pred.) :} & brain & chest & in right & large defect\\
            \hline
            {\bf Comments\hspace{0.21cm}:} & Question demands modality identification but organ is identified and predicted. & Question demands gender identification for which chest analysis is required & Partial prediction. & Type of defect is not identified.\\
            \thickhline
            
            \end{tabular}
            \caption{ Error analysis (Miscellaneous Error): Comparison of Ground Truth answer with the predicted answer for understanding the error types.  }
            \label{table:error-analysis-3}
            \end{adjustwidth}
        \end{table}
     
    \subsection{Error Analysis}
    \label{sec:error-analysis}
        
        On the outputs generated by our model, we conduct thorough error analysis and classify the main sources of errors in the following types:
        
        \begin{enumerate}
            \item {\bf Semantic Error:} This type of error occurs when the system predicts the answer words that are semantically comparable but fails to predict the exact same words as the ground-truth answer.
            
            \item {\bf Modality/Plane Confusion:} This type of error specifically occurs for questions that require identification of the modality/plane. For such questions, the system fails to identify whether only the plane, modality, or a prediction of modality subtype is sufficient, or the question requires a possible combination of these. 
            
            \item {\bf Specification Error:} When the question itself fails to specify how much information is desired in the answer, this type of error occurs where more than one correct answer is possible.
            
            \item {\bf Boundary Loss:} This type of error occurs when the system predicts the correct answer but does not predict the unimportant ground truth answer words that the question itself can determine. 
            
            \item {\bf Miscellaneous Error:} This type of error occurs when the system predicts the information needed to answer the question, but fails to analyze the information collected to answer the question.
        \end{enumerate}
        
        Table~\ref{table:error-analysis-1}, Table \ref{table:error-analysis-2}, and Table \ref{table:error-analysis-3} include a variety of examples along with the justifications to better understand each of these error types. To quantify each error types, we randomly choose $100$ incorrectly generated samples from the CLEF+RAD dataset and categorize them into the five error types. We quantitatively analyze the errors and found that $20.54\%$ errors belong to \textit{Modality/Plane Confusion} and $14.16\%$ errors belong to \textit{Semantic Error}. Similarly, we found $16.48\%$ errors fall into  the \textit{Specification Error}  type. \textit{Boundary Loss} type error contribute to the $20.49\%$ of the total errors. The remaining errors ($28.33\%$) associated with the \textit{Miscellaneous Error} type.
\section{Conclusion}
\label{sec:conclusion}
        In this paper, we propose a hierarchical multi-modal approach to tackle the VQA problem in the medical domain. In particular, we use a question segregation module at the top level of our hierarchy to divide the input questions into two different types (`\textit{Yes/No}' and `\textit{Others}'), followed by individual and independent models at the leaf level, each dedicated to the type of question segregated at the previous level. Our proposed approach can be applied to any related problem where such segregation is possible but it does require non-trivial changes in the architecture. %We use SVM for QS but based on the requirements more rigorous QS techniques can be implemented. 
        To evaluate the usefulness of our proposed model, we conduct experiments on two different datasets, RAD and CLEF18. We also perform experiments on the combined data of the above two datasets to show the generalisability of our approach. Models, when trained with the proposed hierarchy with QS, scored better, outperforming all the stated baseline models. It suggests that questions with different types learn better in isolation having their individual learning paths. Experimental results indicate the effectiveness of our work, depicting its value for the VQA in the medical domain. We also find out that even simple versions of our model are competitive.

        Further analysis of the obtained results reveals that the evaluation metric needs improvement to evaluate VQA in the medical domain. For future work, we plan to investigate a better evaluation strategy for evaluating the task apart from devising a detailed scheme for QS. We also plan to introduce better individual models to handle each of the leaf node tasks.
\section*{Acknowledgment}
Asif Ekbal gratefully acknowledges the Young Faculty Research Fellowship (YFRF) Award supported by the Visvesvaraya PhD scheme for Electronics and IT, Ministry of Electronics and Information Technology (MeitY), Government of India, and implemented by Digital India Corporation (formerly Media Lab Asia).

\bibliography{mybibfile}

\end{document}